\pdfoutput=1
\documentclass[11pt]{article}
\usepackage{acl}
\usepackage{times}
\usepackage{latexsym}
\usepackage[T1]{fontenc}
\usepackage[utf8]{inputenc}
\usepackage{microtype}
\usepackage{graphicx}
\usepackage{afterpage}
\usepackage{float}
\usepackage{arydshln}
\usepackage{multirow} 
\usepackage{booktabs}
\usepackage{colortbl}
\usepackage{xcolor}
\usepackage{amsmath}
\usepackage{amsfonts}  
\usepackage{amssymb}
\usepackage{algorithm}
\usepackage{algorithmic}
\usepackage{pifont}
\usepackage{graphicx}  
\usepackage{subcaption}
\usepackage{placeins}
\usepackage{xurl}
\usepackage{listings}
\usepackage{bm}

\newcommand{\method}{Target-DPO}

\newcommand{\ie}{\textit{i.e.}}

\definecolor{diffadd}{RGB}{220,255,220}
\definecolor{diffdel}{RGB}{255,220,220}

\makeatletter  
\renewcommand\@fnsymbol[1]{%
  \ensuremath{%
    \ifcase#1\or %
    \star\or     %
    \diamond\or   %
    \dagger\or  %
    \mathsection\or %
    \mathparagraph\or %
    \|\or        %
    **\or        %
    \dagger\dagger\or %
    \ddagger\ddagger %
    \else \@ctrerr \fi%
  }%
}  
\makeatother

\title{Teaching Your Models to Understand Code via Focal Preference Alignment}

\author{%
  \textbf{Jie Wu}\thanks{Equal contribution. Work done during the internships of Jie Wu, Haoling Li, and Jianwen Luo at Microsoft Research. Email: \{wujie24,li-hl23\}@mails.tsinghua.edu.cn.}$^{\phi}$,
  \textbf{Haoling Li}$^{\star\phi}$,
  \textbf{Xin Zhang}\thanks{Project leader.}$^{\star\pi}$,
  \textbf{Xiao Liu}$^{\pi}$,
  \textbf{Yangyu Huang}$^{\pi}$,
  \textbf{Jianwen Luo}$^{\sigma}$, \\
  \textbf{Yizhen Zhang}$^{\phi}$,
  \textbf{Zuchao Li}$^{\diamond}$,
  \textbf{Ruihang Chu}\thanks{Corresponding author.}$^{\phi}$,
  \textbf{Yujiu Yang}$^{\phi}$,
  \textbf{Scarlett Li}$^{\pi}$\\
  $^{\phi}$Tsinghua University \quad $^{\pi}$Microsoft Research \quad 
  $^{\sigma}$CASIA \quad
  $^{\diamond}$Wuhan University\\
}

\begin{document}
\maketitle
\begin{abstract}

Preference learning extends the performance of Code LLMs beyond traditional supervised fine-tuning by leveraging relative quality comparisons. In existing approaches, a set of n candidate solutions is evaluated based on test case success rates, with the candidate demonstrating a higher pass rate being labeled as positive and its counterpart with a lower pass rate as negative.
However, because this approach aligns entire failing code blocks rather than pinpointing specific errors, it lacks the granularity necessary to capture meaningful error-correction relationships. As a result, the model is unable to learn more informative error-correction patterns.
To address these issues, we propose \method, a new preference alignment framework that mimics human iterative debugging to refine Code LLMs. \method~explicitly locates error regions and aligns the corresponding tokens via a tailored DPO algorithm. To facilitate it, we introduce the CodeFlow dataset, where samples are iteratively refined until passing tests, with modifications capturing error corrections.
Extensive experiments show that a diverse suite of Code LLMs equipped with \method~achieves significant performance gains in code generation and improves on challenging tasks like BigCodeBench. 
In-depth analysis reveals that \method~yields fewer errors.
Code, model and datasets are in: \href{https://github.com/JieWu02/Target-DPO}{https://github.com/JieWu02/Target-DPO.}
\end{abstract}
\section{Introduction}

\begin{figure}[!t]
    \centering
    \includegraphics[width=1.0\linewidth]{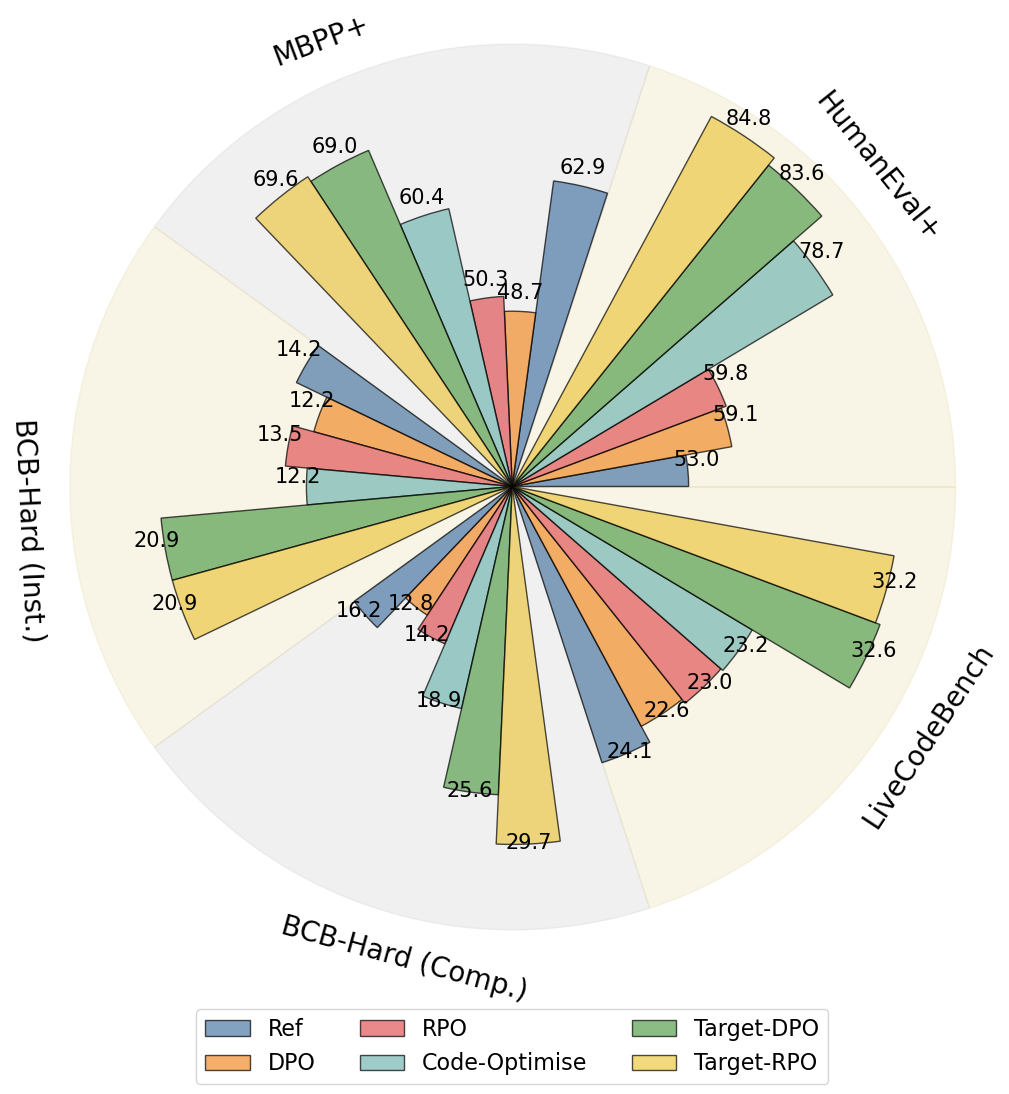}
     \caption{\method~achieves significant performance gains over DPO variants on challenging coding tasks, \ie, BigCodeBench-Hard, with Qwen2.5-Coder-7B.}
    \label{fig:1}
\end{figure}
\begin{figure*}[!t]
    \centering
    \includegraphics[width=1.0\linewidth]{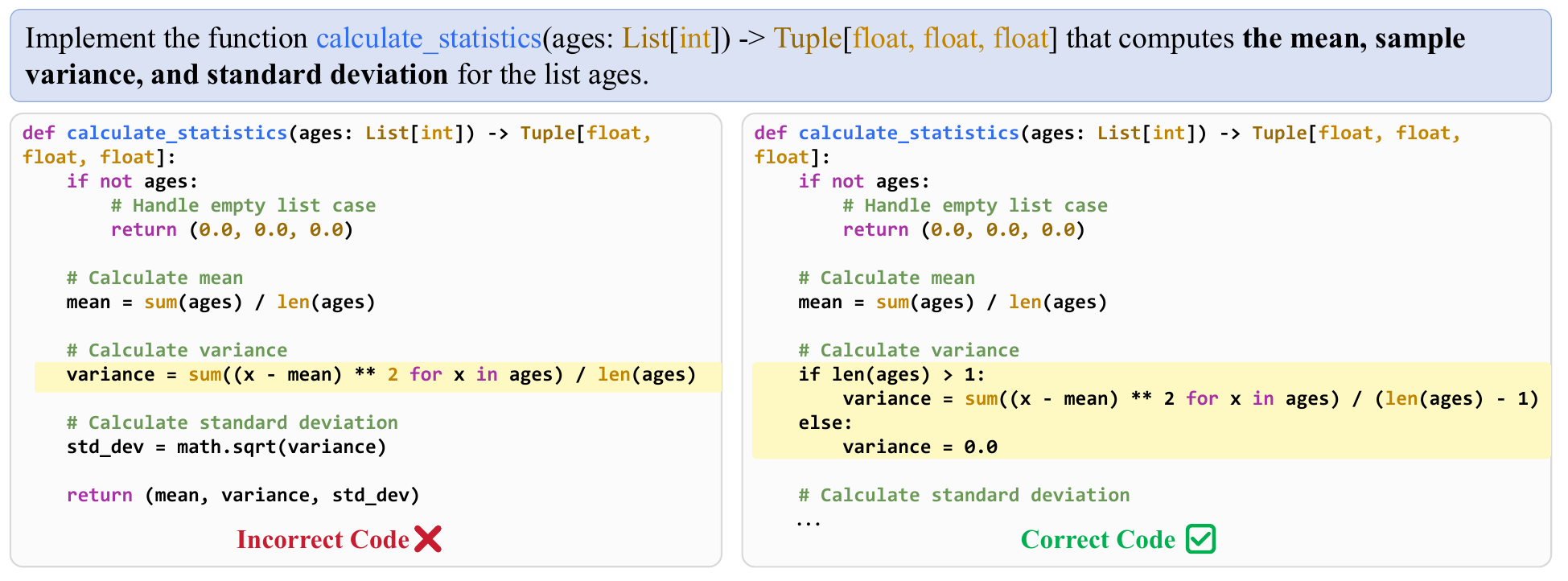}
     \caption{In LLM-generated code, errors are usually confined to critical parts. Minor adjustments to the corresponding erroneous tokens can correct the code while leaving the majority unchanged. Therefore, an effective error correction requires first identifying the key error lines and then performing focal alignment.}
    \label{fig:2}
\end{figure*}

Preference learning offers a promising complement to supervised fine-tuning (SFT)~\cite{DBLP:journals/corr/abs-2308-10792} for improving code generation accuracy in coding large language models (Code LLMs).
Existing methods~\cite{DBLP:journals/corr/abs-2406-06887, zhang2025codedpo,DBLP:journals/corr/abs-2411-13611} mainly rely on unit test feedback to construct preference pairs. In these approaches, a Code LLM generates multiple code snippets as candidates and evaluates each against a suite of test cases. The snippet with the higher pass rate is considered preferred, while the one with the lower pass rate is marked as dispreferred, which forms the pair for preference learning such as Direct Preference Optimization (DPO)~\cite{DBLP:conf/nips/RafailovSMMEF23}. 
However, this paradigm suffers from two critical drawbacks.
First, constructing pairs purely on pass rate cannot guarantee high-quality labels. A high-pass-rate snippet may still carry subtle but crucial bugs, while a low-pass-rate snippet might need only a few modifications to become correct (as shown in Figure~\ref{fig:2}), resulting in noisy preference data.
Second, as errors may be isolated to specific code parts, aligning entire snippets can dilute the correct signal. It forces the model to adjust irrelevant tokens and hinders its ability to learn more specific error patterns~\cite{pal2024smaugfixingfailuremodes,DBLP:journals/corr/abs-2402-05369,DBLP:journals/corr/abs-2407-08639}, which would increase the overfitting risk. The limitations call for a better framework that can pinpoint error regions and apply targeted learning to correct those precise areas.

To tackle these challenges, we draw inspiration from how developers debug code.
Typically, a programmer first locates the module that generates errors based on execution feedback and then focuses on fixing that specific portion until all tests pass. Following this human approach, we introduce \method, a novel framework for preference learning in Code LLMs that leverages iterative debugging insights. Rather than only using pass rate to measure the degree of preference, \method~derives preference pairs from debugging process itself, where the refine steps yield a preference pair, labeling the corrected snippet as preferred and its previous version as dispreferred.
By explicitly contrasting the tokens that resolve the error, 
\method~trains Code LLMs to learn fine-grained alignment for precise error correction, enabling models to truly understand the code.

To support this framework, we efficiently synthesize high-quality preference pairs to create CodeFlow, a novel dataset systematically recording code iterations and corresponding error corrections. Compared to sampling-based methods, CodeFlow enables the efficient creation of preference pairs by (1) generating code snippets and test cases, (2) iteratively refining code until all tests pass, and (3) annotating key token changes between failed and corrected versions. This process ensures that preference learning focuses on the actual error-resolution steps taken by developers.

Building on CodeFlow, we propose an improved DPO algorithm that rewards correct code tokens while penalizing only error-specific tokens in dispreferred samples, minimizing irrelevant noise during preference learning and thus improving efficiency.
Comprehensive ablation studies show how to best select dispreferred samples and how much context to include during alignment, verifying our optimal design for code correction.

We conduct extensive experiments on five public datasets to validate the effectiveness of \method. With only 59k preference pairs, \method~achieves significant performance gains across various base and instruct-tuned Code LLMs. Notably, as shown in Figure~\ref{fig:1}, \method~attains superior results on complex coding tasks like BigCodeBench.
Through detailed ablation studies, we also demonstrate that \method~outperforms alternative strategies by a clear margin.
Our contributions are as follows:
\begin{enumerate}
\item We propose \method, a novel framework that leverages the idea of iterative debugging to address challenges in preference learning, enabling more precise alignment on critical error tokens.
\item We construct a new function-level dataset CodeFlow that iteratively tracks token differences across preference pairs, and propose a tailored adaptation of the DPO algorithm that avoids unnecessary optimization noise.
\item \method~consistently improves performance across diverse benchmarks and various base and instruct-tuned Code LLMs.
\end{enumerate}
\section{\method~Framework}
The \method~framework mimics human iterative debugging to refine Code LLMs. It explicitly identifies error regions and focuses on aligning the corresponding tokens through a tailored DPO algorithm.
To achieve this, \method~follows two steps: (1) synthesizing preference code pairs through an iterative debugging process and locate error regions within code, resulting in the creation of CodeFlow (Section~\ref{sec3.1}), and (2) performing fine-grained and focal alignment by contrasting critical tokens via the designed DPO algorithm (Section~\ref{sec3.2}).
The overall workflow is illustrated in Figure~\ref{fig:4}. In the following sections, we will provide a detailed description of each step.

\subsection{Synthesize Preference Code Snippets}
\label{sec3.1}
As part of our method, we synthesize 59k preference pairs to enable targeted alignment.
In contrast to previous methods that synthesize preference pairs based on pass rate, \method~synthesizes preference code snippets from an iterative debugging process. 
In this process, an initial code snippet is refined until it passes the test cases, and a preference pair is constructed between the final correct version and the previous iteration.

\vspace{1mm}
\noindent \textbf{Generate Raw Code Snippets and Tests.}
To obtain diverse and complex code data, we adopt the practice of EpiCoder~\cite{Wang2025EpiCoderED}, utilizing its feature tree-based synthesis framework with GPT-4o~\cite{openai2024gpt4technicalreport} to generate high-quality code and test cases. This approach directs the LLM to produce a coding task instruction, the corresponding code snippet, and multiple test cases.

To ensure the quality of the generated test cases, we applied several validation measures, including coverage analysis, LLM-based evaluation, and human verification. Detailed discussions and prompting examples are provided in Appendix~\ref{subsec:appendix_synthetic_test_generation}.

\begin{figure*}
    \centering
    \includegraphics[width=1\linewidth]{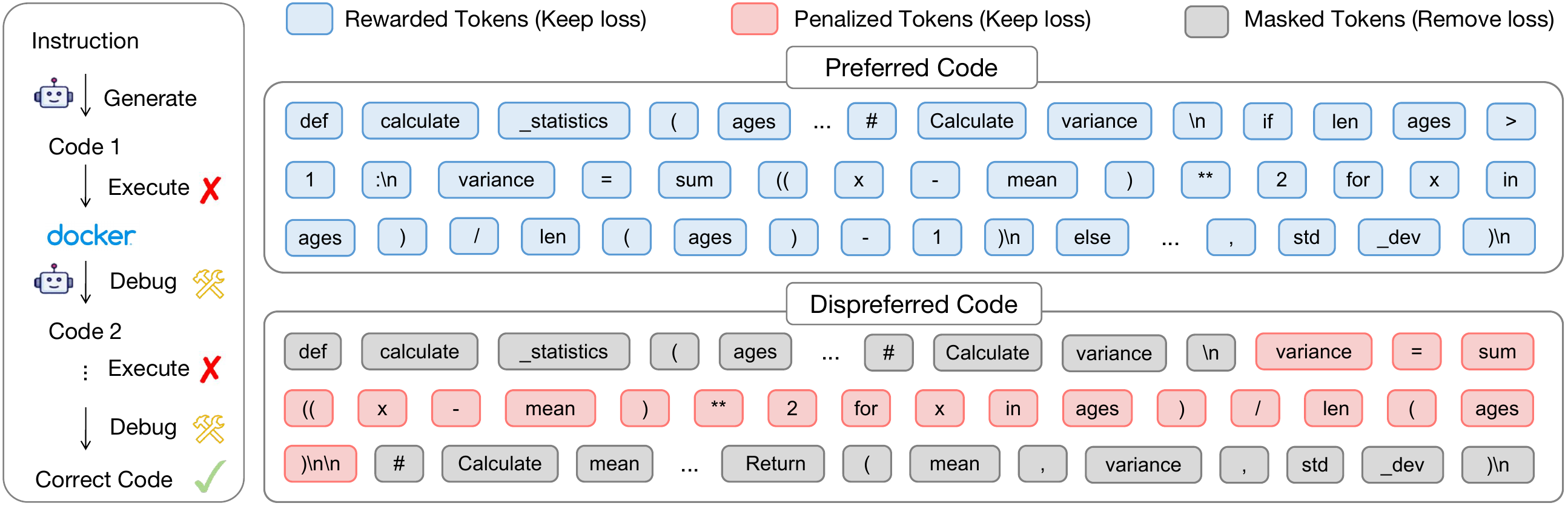}
    \caption{\textbf{Method Overview.} \method~constructs preference pairs via iterative debugging, treating the correct version as preferred and the previous as dispreferred. DPO adaptations enable code LLMs to learn the correct pattern from the preferred code while highlighting critical tokens with a masking strategy in the dispreferred sample.}
    \label{fig:4}
\end{figure*}

\vspace{1mm}
\noindent \textbf{Iterative Refinement via Verification.}
LLMs cannot guarantee the correctness of generated code~\cite{ma2025dynamicscalingunittests}. Therefore, we verify each code sample and refine it through iterative debugging based on execution feedback from verified test cases. As shown in Figure~\ref{fig:4}, when the initial code fails the unit tests, we collect the error information and refine the code iteratively until it passes the test.
The pass rate at the $T$-th iteration is reported in Figure~\ref{fig:dataset_pass_rate} of Appendix~\ref{subsec:appendix_core_method_preference_pairs}.

Our goal is to collect program changes made during the iterative debugging process. To reduce costs, we discard code that fails to be corrected within five iterations. Although this filter removes some extremely challenging cases, it does not make the generated dataset predominantly easier, as clarified in Appendix~\ref{subsec:appendix_core_method_preference_pairs}.
Samples requiring more than five iterations for a solution generally fail to pass all test cases anyway, regardless of additional sampling efforts.

After iterative debugging, we treat the final correct code as the preferred sample and randomly select an earlier version as the dispreferred sample, forming the pair (\(y^+\), \(y^-\)) for preference learning.

\begin{algorithm}[!t]
\caption{Extracting Code Difference}
\label{alg.difference}
\begin{algorithmic}[1]
\REQUIRE Code pair \( y^+ \) and \( y^- \)
\ENSURE Difference lines \( \mathcal{D}^+ \) and \( \mathcal{D}^- \) for \( y^- \)

\STATE Split \( y^+ \) and \( y^- \) into lines: \( y^+_{\text{lines}} \) and \( y^-_{\text{lines}} \)
\STATE Find the LCS of lines between \( y^+_{\text{lines}} \) and \( y^-_{\text{lines}} \)
\STATE Initialize \( \mathcal{D}^+ = \emptyset \) and \( \mathcal{D}^- = \emptyset \)
\STATE \( \mathcal{D}^+ = \{ l^+ \in y^+_{\text{lines}} \mid l^+ \notin \text{LCS} \} \)
\STATE \( \mathcal{D}^- = \{ l^- \in y^-_{\text{lines}} \mid l^- \notin \text{LCS} \} \)
\RETURN \( \mathcal{D}^+ \) and \( \mathcal{D}^- \)
\end{algorithmic}
\end{algorithm}

\vspace{1mm}
\noindent \textbf{Critical Difference Extraction.}
To identify the targeted regions for alignment, we extract the critical differences responsible for the functional divergence between each ($y^+$, $y^-$) pair.
Specifically, we pinpoint the sets of differing lines, $D^+$ and $D^-$, by computing the Longest Common Subsequence (LCS).
Lines not part of the LCS are considered difference lines, as detailed in Algorithm~\ref{alg.difference}.
Consequently, the key modifications distinguishing the preferred from the dispreferred samples are localized within $D^+$ and $D^-$.
These segments encapsulate the changes driving the functional distinctions and represent the critical regions that \method~aims to contrast and align.

\vspace{1mm}
\noindent \textbf{Quality Control for Preference Pairs.}
We initially synthesize 104k instruction data points through iterative refinement. To further ensure data quality, we implement several filtering measures. Specifically, rule-based filtering is applied to remove trivial or uninformative samples, such as those where: (i) \( D^- \) consists only of comments; (ii) \( D^- \) exceeds 20 lines; (iii) \( y^+ \) or \( y^- \) exceeds 2048 tokens; and (iv) Samples where the abstract syntax tree (AST) of \( y^+ \) and \( y^- \) are identical.

The application of these filters reduced the dataset to 84k samples. In the next stage, we utilized GPT-4o as an LLM-judge to assess whether a significant logical distinction existed between \( y^+ \) and \( y^- \). We further filter out pairs where the differences are limited to code formatting, comments, variable names, whitespace, or blank lines. These efforts ensure that the selected and rejected samples reflect key functional differences. The final dataset, consisting of 59k samples, is thus prepared for preference-based alignment training.

\subsection{Targeted Preference Alignment}
\label{sec3.2}
Direct Preference Optimization (DPO) directly optimizes the policy model using relative quality comparisons.
Given a prompt \(x\), a preference pair \((y^{+}, y^{-})\), where \(y^+\) is of higher quality than \(y^{-}\),
DPO aims to maximize the probability of the preferred response \(y^{+}\) while minimizing that of the less desirable response \(y^{-}\). The KL divergences for \(y^{+}\) and \(y^{-}\) are defined as:
\begin{equation}
\scalebox{0.9}{$
\mathcal{K}^{+} = \log \frac{\pi_\theta(y^{+}|x)}{\pi_{\text{ref}}(y^+|x)}, \; \mathcal{K}^{-} = \log \frac{\pi_\theta(y^{-}|x)}{\pi_{\text{ref}}(y^-|x)},
$}
\end{equation}
and the optimization objective \(\mathcal{L}_{\text{DPO}}(\pi_\theta; \pi_{\text{ref}})\) is:
\begin{equation}
\scalebox{0.9}{$
    \mathcal{L}_{\text{DPO}}=-\mathbb{E}_{(x, y^+, y^-) \sim \mathcal{D}} \left[ \log \sigma \left( \beta \left( \mathcal{K}^+ - \mathcal{K}^- \right) \right) \right]
$}
    \label{DPO_eq}
\end{equation}
DPO optimizes the expectation over the pairwise preference dataset \(\mathcal{D}\), and \(\sigma\) is the sigmoid function.

While Direct Preference Optimization (DPO) has demonstrated effectiveness in domains such as mathematics~\cite{lai2024stepdpostepwisepreferenceoptimization}, its standard objective function, as shown in Equation~\eqref{DPO_eq}, may be suboptimal for preference-based alignment in code generation, as a large portion of the tokens in \(y^+\) and \(y^-\) are identical, with only minor differences. This can confuse the policy model in identifying the critical differences necessary for functional correctness, and diminish alignment gains~\cite{pal2024smaugfixingfailuremodes,DBLP:journals/corr/abs-2402-05369,DBLP:journals/corr/abs-2407-08639}.

To help code LLMs better grasp the critical tokens driving functional differences between preference pairs, we modify the DPO algorithm to highlight key tokens in the dispreferred code snippet using a masking strategy.
Specifically, given $y^-=[y_1^-, y_2^-, ..,y_L^-]$ containing $L$ tokens, vanilla DPO computes $\mathcal{K}^-$ as:
\begin{equation}
\scalebox{0.9}{$
    \begin{aligned}
    K^- = \log \frac{\pi_\theta(y^-|x)}{\pi_{\text{ref}}(y^-|x)}
        &= \log  \frac{\prod_{i=1}^{L} \pi_{\theta}(y_i^- | x)}{\prod_{i=1}^{L} \pi_{\text{ref}}(y_i^- | x)} \\
        &= \sum_{i=1}^{L} \log  \frac{\pi_{\theta}(y_i^- | x)}{\pi_{\text{ref}}(y_i^- | x)}
    \end{aligned}
$}
\end{equation}

We make the following adaptations to $\mathcal{K}^-$ while keeping $\mathcal{K}^+$ unchanged:
\begin{equation}
\scalebox{0.9}{$
K^{+'} = K^+
$}
\label{k_w'}
\end{equation}
\begin{equation}
\scalebox{0.9}{$
K^{-'} = \sum_{i=1}^{L} \mathbb{I}(y_i^- \in \text{D}^-) \log  \frac{\pi_{\theta}(y_i^- | x)}{\pi_{\text{ref}}(y_i^- | x)}
$}
\label{k_l'}
\end{equation}

Equation~\eqref{k_w'} guides the code LLM to learn correct code generation patterns from $y^+$.
In contrast, Equation~\eqref{k_l'} explicitly focuses on contrasting critical tokens within $y^-$.
It achieves this by masking tokens in the dispreferred code that do not appear in $D^-$, thereby excluding correct tokens in $y^-$ from the loss computation, as illustrated on the right side of Figure~\ref{fig:4}.
By penalizing critical tokens that cause functional errors and preventing over-optimization on tokens common to both \(y^+\) and \(y^-\), \method~achieves a more fine-grained alignment tailored for code, improving upon previous sample-level optimization approaches. This refined strategy enables code LLMs to better internalize correct coding patterns and more effectively identify crucial token-level errors.

Our loss also targets pairwise optimization:
\begin{equation}
\scalebox{0.9}{$
    \mathcal{L}^{'}_{\text{DPO}}=-\mathbb{E}_{(x, y^+, y^-) \sim \mathcal{D}} \log \sigma \left( \beta \left( \mathcal{K}^{+'}-\mathcal{K}^{-'} \right) \right) 
$}
    \label{DPO_eq_ours}
\end{equation}
Correspondingly, the RPO loss~\cite{liu2024provably,pang2024iterative}, a variant of DPO, consists of a weighted SFT loss on \(y^+\), scaled by \(\alpha\). Our modified DPO loss also complements RPO, and the RPO-format \(\mathcal{L}^{'}_{\text{RPO}}\) loss is:
\begin{equation}
\scalebox{0.9}{$
    \mathcal{L}_{\text{SFT}} = -\mathbb{E}_{(x, y^+) \sim \mathcal{D}} \left[ \log p_{\theta}(y^+ | x) \right]
$}
\end{equation}
\begin{equation}
\scalebox{0.9}{$
    \mathcal{L}_{\text{RPO}^{'}} = \mathcal{L}^{'}_{\text{DPO}} + \alpha \mathcal{L}_{\text{SFT}}
$}
\end{equation}

\begin{table*}[!t]
\centering
\small
\begin{tabular}{@{}lp{1.95cm}cccccccccc@{}}
\toprule
\multirow{2}{*}{\centering \textbf{Model}} & \multirow{2}{*}{\centering \textbf{Variant}} & 
\multicolumn{2}{c}{\textbf{HumanEval}} & 
\multicolumn{2}{c}{\textbf{MBPP}} & 
\multicolumn{2}{c}{\textbf{BCB-Full}} & 
\multicolumn{2}{c}{\textbf{BCB-Hard}} & 
\multicolumn{1}{c}{\textbf{LCB}} &
\multirow{2}{*}{\centering \textbf{Avg.}} \\
& & \textit{Base} & \textit{Plus} & \textit{Base} & \textit{Plus} & \textit{Comp.} & \textit{Inst.} & \textit{Comp.} & \textit{Inst.} & \textit{Inst.} & \\
\midrule
\multirow{7}{*}{DS-Coder-7B-Ins-v1.5}
& Ref. & 75.6 & 71.3 & 75.2 & 62.3 & 43.8 & 35.5 & 15.5 & 10.1 & 20.6 & 45.5 \\
& DPO & 69.5 & 65.2 & 77.2 & 67.2 & 46.1 & 37.9 & 12.2 & 14.2 & 20.4 & 45.5 \\
& RPO & 65.2 & 59.8 & 75.7 & 66.1 & 43.2 & 37.5 & 10.8 & 13.8 & 20.2 & 43.6 \\
& Code-Optimise & 64.6 & 60.4  & 78.8 & \textbf{69.3} & 45.2 & 36.5 & 13.5 & 13.5 & 21.3 & 44.8 \\
& \method & 76.2 & 72.0  & \textbf{79.1} & 65.3 & 47.5 & 37.8 & \textbf{22.3} & 17.6 & 21.8 & 48.8 \\
& Target-RPO & \textbf{78.0} & \textbf{73.2} & 78.8 & 67.2 & \textbf{49.3} & \textbf{39.0} & 20.9 & \textbf{20.9} & \textbf{22.0} & \textbf{49.9} \\
\midrule
\multirow{7}{*}{CodeQwen1.5-7B-Chat} 
& Ref. & 83.5 & 78.7 & 79.4 & 69.0 & 43.6 & 39.6 & 15.5 & \textbf{18.9} & 15.3 & 49.3 \\
& DPO & 79.3 & 73.8 & 79.9 & 69.0 & 43.3 & 36.1 & 14.9 & 10.8 & 15.5 & 47.0 \\
& RPO & 79.3 & 73.2  & 80.2 & 68.8 & 41.6 & 32.5 & 14.8 & 10.6 & 12.9 & 46.0 \\
& Code-Optimise & 78.5 & 75.0 & 80.7 & 69.6 & 43.3 & 36.1 & 17.6 & 11.5 & 16.2 & 47.6 \\
& \method & 89.6 & 85.4 & \textbf{83.9} & 69.8 & \textbf{48.7} & \textbf{39.9} & 20.3 & 16.9 & \textbf{18.1} & \textbf{52.5} \\
& Target-RPO & \textbf{89.6} & \textbf{86.0} & 82.5 & \textbf{70.4} & 48.4 & 38.3 & \textbf{20.3} & 18.2 & 17.2 & 52.3 \\
\midrule
\multirow{6}{*}{StarCoder2-15B}
& Ref. & 46.3 & 37.8 & 66.2 & 53.1 & 38.4 & - & 12.2 & - & - & - \\
& DPO & 51.8 & 45.1 & 63.8 & 42.9 & 27.3 & 16.2 & 8.1  & 5.4 & 12.7 & 30.4 \\
& RPO & 53.0 & 45.7 & 63.0 & 42.6 & 28.7 & 17.2 & 9.1 & 6.0 & 13.1 & 30.9 \\
& Code-Optimise & 61.0 & 54.9 & 66.5 & 53.4 & 31.8 & 18.8 & 6.8 & 6.1 & 14.9 & 34.9 \\
& \method & 70.7 & 64.6 & \textbf{67.2} & \textbf{54.5} & 39.7 & 37.7  & 17.6 & 16.9 & 18.7 & 43.1 \\
& Target-RPO & \textbf{73.2} & \textbf{65.2} & 65.9 & 53.4 & \textbf{40.3} & \textbf{38.8} & \textbf{18.9} & \textbf{18.2} & \textbf{19.4} & \textbf{43.7} \\
\midrule
\multirow{6}{*}{Qwen2.5-Coder-7B}
& Ref. & 61.6 & 53.0 & 76.9 & 62.9 & 45.8 & 40.2 & 16.2 & 14.2 & 24.1 & 43.9 \\ 
& DPO & 71.3 & 59.1 & 76.2 & 48.7 & 38.8 & 28.5 & 12.8  & 12.2 & 22.6 & 41.1 \\
& RPO & 71.3 & 59.8 & 70.9 & 50.3 & 39.8 & 29.7 & 14.2 & 13.5 & 23.0 & 41.4 \\
& Code-Optimise & 82.3 & 78.7 & 76.2 & 60.4 & 48.5 & 39.6  & 18.9  & 12.2 & 23.2 & 48.9 \\
& \method & 89.0 & 83.6 & 83.1 & 69.0 & 52.7 & 41.0 & 25.6 & 20.9 & 32.6 & 55.3 \\
& Target-RPO & \textbf{89.6} & \textbf{84.8} & \textbf{83.3} & \textbf{69.6} & \textbf{53.3} & \textbf{43.1} & \textbf{29.7} & \textbf{20.9} & \textbf{32.2} & \textbf{56.3} \\
\bottomrule
\end{tabular}
\caption{Pass@1 (\%) results of different LLMs on HumanEval, MBPP, BigCodeBench, and LiveCodeBench-v5 (LCB) under greedy
decoding setting. We conducted the evaluation on the Full and Hard subsets of BigCodeBench (BCB), including the Complete
(Comp.) and Instruct (Inst.) tasks. The best results are highlighted in Bold.}
\label{tab:performance_comparison}
\end{table*}
\section{Experiments}
\noindent
\textbf{Experiment Setup.}
For our \method, the learning rate is set to 1e-5 for the 7B code LLMs and 5e-6 for the 15B models, using a global batch size of 128, with a cosine scheduler and warm-up.
The maximum sequence length is set to 2048 tokens. Detailed training settings are presented in Appendix~\ref{subsec:appendix_training_inference_params}
For the DPO algorithm, $\beta$ is set to 0.1, and for RPO, $\alpha$ is set to 1.0.
The rationale behind the choice of $\alpha$ and $\beta$ is supported by ablation studies presented in Appendix~\ref{subsec:appendix_ablations_beta_alpha_results}.
$\pi_\theta$ and $\pi_{\text{ref}}$ are both initialized with the weights of the evaluated model, while $\pi_{\text{ref}}$ keeps frozen during training.

\noindent
\paragraph{Benchmarks.}
We evaluate the Code LLMs using multiple benchmarks: \textsc{HumanEval} Base~\cite{chen2021evaluatinglargelanguagemodels}, \textsc{HumanEval} Plus~\cite{10.5555/3666122.3667065}, Mostly Basic Python Problems (\textsc{MBPP} Base~\cite{austin2021programsynthesislargelanguage}, \textsc{MBPP} Plus), LiveCodeBench (LCB)~\cite{jain2024livecodebenchholisticcontaminationfree} (v5 with problems released between May 2023 and Jan 2025), and \textsc{Big-CodeBench} (BCB)~\cite{zhuo2025bigcodebench} with instruct and completion splits.
We report the pass@1 score under greedy decoding.

\noindent
\paragraph{Evaluated Models and Baselines.} We evaluate models including DeepSeek-Coder-7B-Instruct-v1.5~\cite{guo2024deepseekcoderlargelanguagemodel}, CodeQwen1.5-7B-Chat~\cite{qwen}, as well as base models such as Qwen2.5-Coder-7B~\cite{hui2024qwen2} and StarCoder2-15B~\cite{lozhkov2024starcoder2stackv2}.
Results for the 32B model are in Appendix~\ref{subsec:appendix_scaling_model_size_results}. CodeDPO and PLUM are compared using their reported results, as their data and code are currently unavailable. Code-Optimise~\cite{gee2025codeoptimiseselfgeneratedpreferencedata} is reproduced using GPT-4o, with 100 solutions sampled at a temperature of 0.6 for each problem. The DPO-PvF setting results for Code-Optimise are reported.

\section{Main Results}
Table~\ref{tab:performance_comparison} presents a comparison between baseline models, DPO variants, and \method. We discuss the findings from the following perspectives.

\vspace{1mm}
\noindent
\textbf{Focal Alignment Outperforms Global Alignment.}
Preference pairs from iterative debugging differ significantly from those in datasets like Code-Optimise, causing a performance drop with vanilla DPO or RPO. While some settings, like DS-Coder-7B-Instruct-DPO, show gains on \textsc{BigCodeBench}, DPO and RPO generally underperform compared to baselines.
In typical correction scenarios, an LLM modifies only a small portion of the code to fix errors, creating highly similar preference pairs. This overlap introduces ambiguity, as identical tokens in both positive and negative examples weaken the model's ability to distinguish meaningful differences.

This degradation underscores the need for explicit mechanisms to focus the policy model on tokens responsible for functional faults. \method~addresses this by emphasizing error tokens in the dispreferred code and explicitly contrasting the critical edits. 
As shown in Table\ref{tab:performance_comparison}, \method~outperforms DPO by 3.3\% and 5.9\% on average across benchmarks, while Target-RPO yields improvements ranging from 6.3\% to 12.4\%.

\begin{figure}[!b]
    \centering
    \includegraphics[width=1.\linewidth]{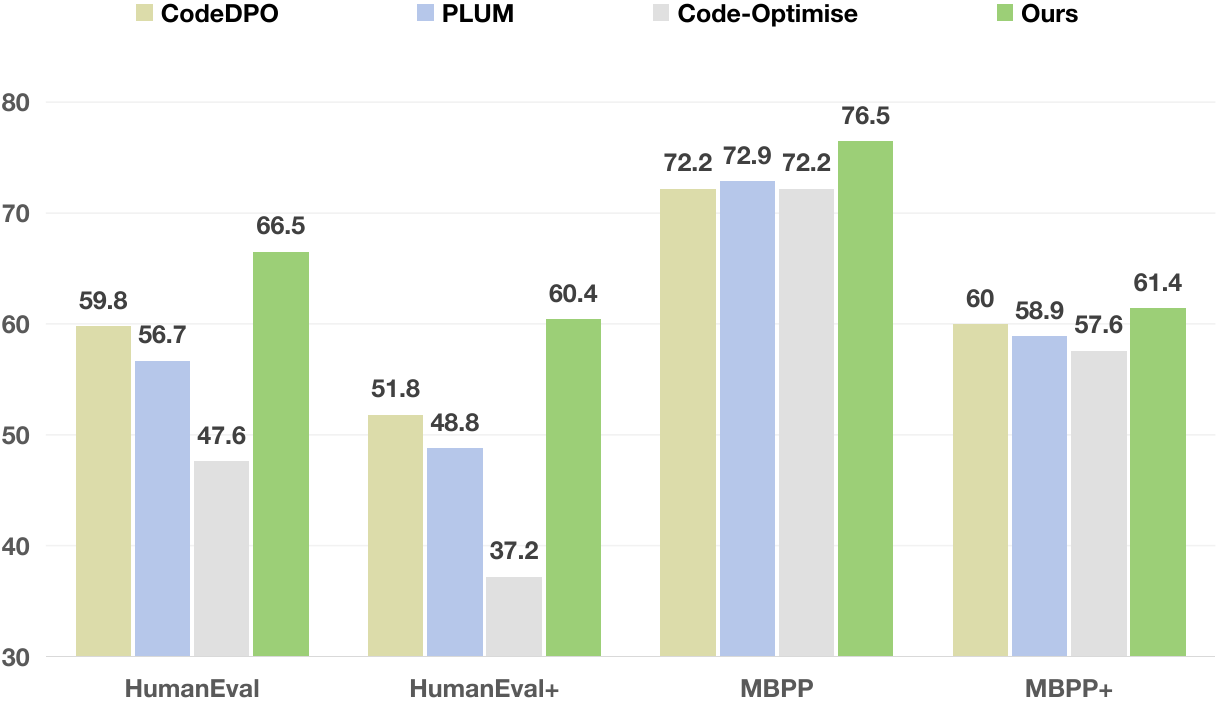}
    \caption{Comparison with CodeDPO, PLUM, and Code-Optimise using DeepSeekCoder-6.7B. Additional results are provided in Appendix~\ref{subsec:appendix_comparison_other_methods}.}
    \label{fig:5}
\end{figure}

\vspace{1mm}
\noindent
\textbf{\method~Achieves Significant Improvements over Methods that Rely on Coarse-grained Pass Rate Signals.}
While methods like PLUM and CodeDPO, which construct preference pairs by testing multiple sampled solutions, offer a straightforward and effective approach, their reliance on coarse-grained pass/fail signals inherently limits the model's ability to learn nuanced error correction and generalize improvements. As shown in Figure~\ref{fig:5}, this limitation becomes apparent when compared to our \method.

\noindent
\textbf{\method~Improves Challenging Coding Task.}
We highlight that the \method~framework has the potential to boost Code LLMs to solve complex coding tasks.
Notably, Qwen2.5-Coder-7B equipped with \method~achieves a 29.7\% pass@1 score on BigCodeBench Complete Hard, matching the performance of larger Code LLMs DeepSeek-Coder-V2-Instruct (29.7\%) and Claude-3-Opus (29.7\%)~\cite{anthropic2024claude}, and approaching Llama-3.1-405B-Instruct (30.4\%)~\cite{grattafiori2024llama3herdmodels}. When given more attempts, \method~achieves pass@5 of 45.7\%, outperforming DeepSeek-R1 (40.5\%) ~\cite{deepseekai2025deepseekr1incentivizingreasoningcapability} and GPT-o1 (40.2\%). On the Instruct Hard split, pass@5 of the \method-Qwen is 34.7\%, comparable to the performance of GPT-o3-mini (33.1\%).

\begin{figure*}[!t]
    \centering
    \includegraphics[width=0.9\linewidth]{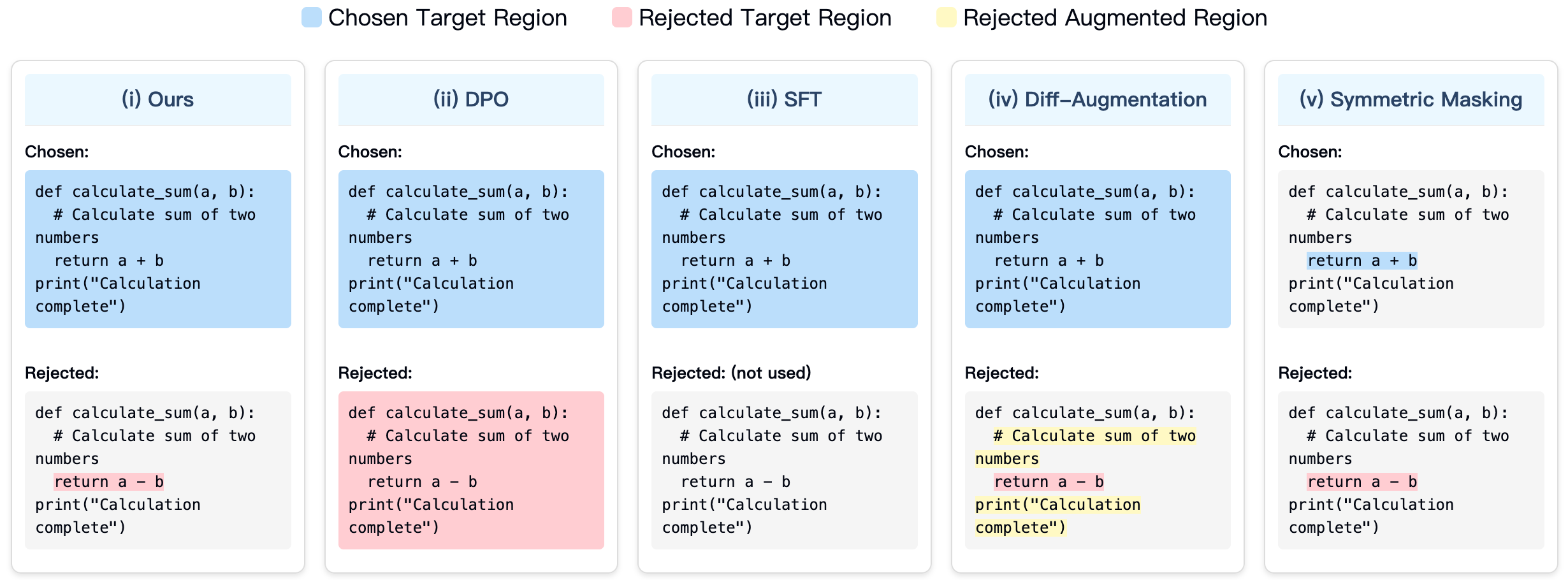}
    \caption{Illustration for Target-DPO and its ablations. Target-DPO rewards correct code tokens while penalizing only error-specific tokens in rejected code, teaching models to truly understand code through targeted alignment.}
    \label{fig:ablations_illustration}
\end{figure*} 

\section{Ablation Study}

\begin{table*}[!htbp]
\centering
\small
\begin{tabular}{lccccc}
\toprule
 & \textbf{HumanEval(Avg)} & \textbf{MBPP(Avg)} & \textbf{BCB(Complete)} & \textbf{BCB(Instruct)} & \textbf{Average} \\
\midrule
SFT (EpiCoder 40k) & 83.9 & 75.5 & 50.9 & 39.1 & 62.4 \\
SFT (EpiCoder 80k) & 85.1 & \textbf{78.9} & 52.3 & 39.4 & 63.9 \\
SFT (EpiCoder 380k) & \underline{85.7} & \underline{77.8} & \textbf{53.4} & \textbf{43.8} & \textbf{65.2} \\
SFT (CodeFlow 59k) & 85.5 & 75.7 & 51.6 & 39.1 & 62.9 \\
Our \method~(59k) & \textbf{87.2} & 76.5 & \underline{53.3} & \underline{43.1} & \underline{65.0} \\
\bottomrule
\end{tabular}
\caption{Results of EpiCoder-SFT with varying amounts of training data and our method on Qwen2.5-Coder-7B.}
\label{tab:epicoder_comparison}
\end{table*}

\begin{table*}[!t]
\centering
\small
\begin{tabular}{lccllccccc}
\toprule
\multirow{2}{*}{} & \multirow{2}{*}{\textbf{Aug}} & \multirow{2}{*}{\textbf{Hybrid}} & \multicolumn{2}{c}{\textbf{HumanEval}} & \multicolumn{2}{c}{\textbf{MBPP}} & \multirow{2}{*}{\textbf{BCB-Inst}} & \multirow{2}{*}{\textbf{BCB-Comp}} & \multirow{2}{*}{\textbf{Average}} \\
\cmidrule(lr){4-5} \cmidrule(lr){6-7} 
& & & \textit{\textbf{Base}} & \textit{\textbf{Plus}} & \textit{\textbf{Base}} & \textit{\textbf{Plus}} & & & \\ 
\midrule
CodeQwen1.5-7B-Chat & - & - & 83.5 & 78.7 & 79.4 & 69.0 & 39.6 & 43.6 & 65.6 \\
\midrule
\multirow{3}{*}{Target-DPO} & \ding{55} & \ding{52} & 83.5 & 79.3 & 81.5 & 66.1 & 38.2 & 44.5 & 65.5 \\
& \ding{52} & \ding{55} & 83.5 & 76.8 & 80.2 & 65.1 & 34.7 & 44.5 & 64.1 \\
& \ding{55} & \ding{55} & \textbf{89.6} & \underline{85.4} & \textbf{83.9} & \underline{69.8} & \textbf{39.9} & \textbf{48.7} & \textbf{69.6} \\
\midrule
\multirow{3}{*}{Target-RPO} & \ding{55} & \ding{52} & 84.8 & 79.3 & 81.5 & 67.7 & 34.6 & 44.0 & 65.3 \\
& \ding{52} & \ding{55} & 86.0 & 79.9 & 81.5 & 65.9 & 36.0 & 45.0 & 65.7 \\
& \ding{55} & \ding{55} & \textbf{89.6} & \textbf{86.0} & \underline{82.5} & \textbf{70.4} & \underline{38.3} & \underline{48.4} & \underline{69.2} \\
\bottomrule
\end{tabular}
\caption{Ablation study on how much contextual information from negative examples is beneficial for \method, evaluated using CodeQwen1.5-7B-Chat. Additional results with Qwen2.5-Coder-7B are provided in Appendix~\ref{sec:appendix_in_depth_ablations}.}
\label{ablation}
\end{table*}

\begin{table*}[!t]
\centering
\small
\begin{tabular}{lccccc}
\toprule
& \textbf{HumanEval(Avg)} & \textbf{MBPP(Avg)} & \textbf{BCB(Complete)} & \textbf{BCB(Instruct)} & \textbf{Avg.} \\
\midrule
SFT (Correct) & \textbf{85.5} & \textbf{75.7} & \textbf{51.6} & \textbf{39.1} & \textbf{63.0} \\
SFT (Incorrect) & 82.7 & 73.5 & 49.2 & 38.6 & 61.0 \\
\bottomrule
\end{tabular}
\caption{Performance comparison of SFT on correct and incorrect code from CodeFlow uisng Qwen2.5-Coder-7B.}
\label{sft_on_pos_neg}
\end{table*}

Despite the effectiveness of \method~in pinpointing critical error regions, there remain open questions about how best to incorporate negative examples and how much context is truly beneficial for code correction. We therefore explore several settings:
(i) \textbf{SFT}: Supervised fine-tuning using the positive sample from the preference pair; (ii) \textbf{Hybrid Training}: Half of the samples in a batch are trained using vanilla DPO, while the other half follows the \method~approach; (iii) \textbf{Diff-Augmentation}: provide more context for the dispreferred sample by including 1 or 2 lines of tokens before and after $D^-$; and (iv) \textbf{Symmetric Masking Strategy}: The Code LLMs learn from the tokens in $D^+$ rather than the full sequence of positive sample.
In Figure~\ref{fig:ablations_illustration}, we illustrate these settings.

\noindent
\textbf{Supervised Fine-Tuning.}
A comparison with EpiCoder-SFT, considering varying amounts of training data, is shown in Table \ref{tab:epicoder_comparison}. 
Our \method~achieves performance comparable to the strong SFT baseline EpiCoder-380k using only 59k training samples (about one-sixth), unveiling the power of targeted alignment.

The positive samples undergo iterative debugging and are validated by test cases, they maintain high quality, allowing SFT to achieve reasonably strong performance.
However, SFT overlooks dispreferred samples, missing the opportunity to contrast and precisely align positive and negative examples.
In contrast, \method~not only leverages error-free code to increase the likelihood of correct code but also precisely penalizes tokens responsible for critical errors, achieving finer-grained alignment and better performance consequently.

\noindent
\textbf{Hybrid Training \& Diff-Augmentation.} Both settings expose Code LLMs to more tokens from the dispreferred samples but differ in scope: in Hybrid Training, 50\% of the training samples use the entire dispreferred sequence, while Diff-Augmentation provides a small token window around the \( D^- \). Table~\ref{ablation} shows that while adding extra context around \( D^- \) may appear beneficial, it often introduces noise that confuses the model, making it unclear which parts need local alignment, ultimately leading to degraded performance.

We find that concentrating solely on the most critical tokens yields better results, highlighting the importance of accurately grounding these tokens for more effective targeted alignment. The iterative debugging process naturally supports this precise localization, as typically only a small portion of the code changes between iterations, while the majority remains unchanged. These targeted regions can be easily identified using the Longest Common Subsequence (LCS), allowing meaningful differences to be isolated with high precision.

\noindent
\textbf{Symmetric Masking Strategy.} When training with the symmetric masking, where Code LLMs learn from both \(D^+\) and \(D^-\) without access to the full positive sample, the model struggles to retain its core code generation capabilities and fails to benchmark effectively. The primary goal of Code LLMs is to generate complete and correct code. Although learning symmetrically from both \(D^+\) and \(D^-\) may seem appealing, the focus should be on ensuring Code LLMs learn from fully correct code rather than fragmented pieces. Without complete code contexts, the positive sample cannot properly align with the instruction, leading to incomplete and misleading signals in the learning process.

\noindent \textbf{Generated Test Cases can Distinguish Good and Error Code.}
Table~\ref{sft_on_pos_neg} compares supervised fine-tuning using either preferred or dispreferred samples. The results show that SFT on preferred samples outperforms that on dispreferred ones by an average of 2.0\%. This quality gap between pairs, introduced through debugging iterations, suggests that test cases effectively differentiate high- and low-quality code snippets, providing training pairs with clear quality contrast. In our debugging pattern, the quality differences between code snippets are primarily influenced by the feedback from test cases, indicating that test cases can reliably distinguish between good and bad code when verified through dedicated efforts.
\section{Related Work}
\vspace{-0.5mm}
\noindent
\textbf{Code Language Models.}
Powerful Code LLMs like Qwen2.5-Coder~\cite{hui2024qwen25codertechnicalreport}, DeepSeek-Coder~\cite{guo2024deepseekcoderlargelanguagemodel}, StarCoder~\cite{li2023starcodersourceyou,lozhkov2024starcoder2stackv2}, Magicoder~\cite{10.5555/3692070.3694228} and EpiCoder~\cite{wang2025epicoder}  demonstrate their capabilities in various code generation tasks.
Current Code LLMs primarily focus on supervised fine-tuning during the post-training stage. While SFT enables Code LLMs to learn the correct patterns, it fails to effectively make them aware of incorrect patterns or how to rectify errors in code. In this work, \method~framework aims to enable Code LLMs to further learn through pairwise contrasting of critical tokens~\cite{lin2024critical}, allowing Code LLMs to continually improve.

\noindent
\textbf{Reinforcement Learning} (RL)~\cite{hu2025coarsetofineprocessrewardmodeling,kaufmann2024surveyreinforcementlearninghuman} maximizes the following objective for a prompt $x$ and response $y$:
\[
\max_{\pi_\theta} \mathbb{E}_{x \sim D_p, y \sim \pi_\theta(\cdot|x)} \left[ r(x, y) - \beta \log \frac{\pi_\theta(y|x)}{\pi_{\text{ref}}(y|x)} \right]
\]
where $D_p$ is the dataset, $\pi_\theta$ is the policy model to be optimized, $\pi_{\text{ref}}$ is the reference, and $\beta$ controls the degree of regularization.
RL for code generation attracts attention recently~\cite{dou-etal-2024-stepcoder,10.1145/3675395,sun2024surveyreasoningfoundationmodels,miao2024aligningcodellmsdirectpreference,dai2025process}.
A commonly used approach is DPO~\cite{DBLP:conf/nips/RafailovSMMEF23}, which eliminates the need for an explicit reward model $r$. Variants like RPO~\cite{liu2024provably,pang2024iterative} and KTO~\cite{10.5555/3692070.3692574} are also frequently used in optimizing code generation.

\noindent
\textbf{Preference Pair Construction.} 
Existing methods construct preference pairs by ranking candidate solutions based on pass rates. PLUM~\cite{DBLP:journals/corr/abs-2406-06887} constructs preference pairs by ranking candidate code solutions based on passed test cases. Code-Optimise~\cite{gee2025codeoptimiseselfgeneratedpreferencedata} incorporates efficiency as an additional learning signal, augmented with annotations from unit test feedback and execution time. 
AceCoder~\cite{10.1145/3675395} selects pairs with distinct pass rate differences. 
DSTC~\cite{liu2024dstcdirectpreferencelearning} constructs preference pairs using self-generated code and tests.
A related concurrent work is CodeDPO, which formulates preference learning as a direct optimization problem using pass/fail signals and proposes a PageRank-inspired algorithm to select high-quality preference pairs. 
In contrast, our method aligns code LLMs through error-resolving edits rather than relying on coarse-grained execution outcomes. This fine-grained supervision provides richer training signals that better capture the semantics of code correction, resulting in improved performance, as verified in Figure~\ref{fig:5}.
\section{Conclusion}
We present \method, a novel preference alignment framework that emulates human iterative debugging to capture critical errors in incorrect code for precise optimization. \method~first identifies error-prone regions and applies an improved DPO algorithm contrasting pivotal segments, teaching Code LLMs to understand and correct code through targeted preference alignment, achieving promising
coding performance.
To support this framework, we efficiently synthesize high-quality preference pairs to create CodeFlow, where each sample undergoes iterative refinement until it passes unit tests, with the modification history providing a natural record of error corrections. Extensive experiments show that \method-equipped Code LLMs achieve significant performance improvements in code generation and excel in tackling basic and complex coding tasks.
\section*{Limitations}
\method~is inspired by the debugging pattern of developers, serving as a novel framework for fine-grained preference learning in Code LLMs. Instead of using pass rate alone, \method~derives preference pairs from iterative debugging process. By contrasting critical tokens between a corrected version and its preceding iteration, \method~helps the model to understand code through targeted alignment.
However, this study focuses on a dataset of 59k samples without further expansion, which may limit generalizability, but offers opportunities for future exploration with larger data.

\section*{Acknowledgments}
This work was partly supported by the National Natural Science Foundation of China (Grant No. 62576191), the research grant No. CT20240905126002 of the Doubao Large Model Fund, and the National Natural Science Foundation of China (No. 62306216).
\bibliography{ref}
\clearpage
\newpage

\appendix
\section*{Appendix}
In this appendix, we first provide more details of our core methodology, including preference pair construction and implementation specifics (Section~\ref{sec:appendix_methodology_data_construction}). Section~\ref{sec:appendix_experimental_results_comparisons} then introduces the evaluation benchmarks and presents comprehensive experimental results, showcasing performance on various benchmarks and detailed comparisons against EpiCoder and other relevant methods. Subsequently, we provide in-depth analyses such as scaling laws, ablation studies on key parameters, data diversity assessments, error pattern examinations, and efficiency evaluations (Section~\ref{sec:appendix_in_depth_ablations}).

\section{Methodology and Data Construction}
\label{sec:appendix_methodology_data_construction}

This section details the core methodology of our proposed approach, including the iterative refinement process for preference pair construction and the generation and quality assessment of synthetic test data. Implementation specifics relevant to these methodological aspects are also covered.

\subsection{Iterative Refinement and Preference Pair Construction}
\label{subsec:appendix_core_method_preference_pairs}
\begin{figure}[!b]
    \centering
    \includegraphics[width=1.0\linewidth]{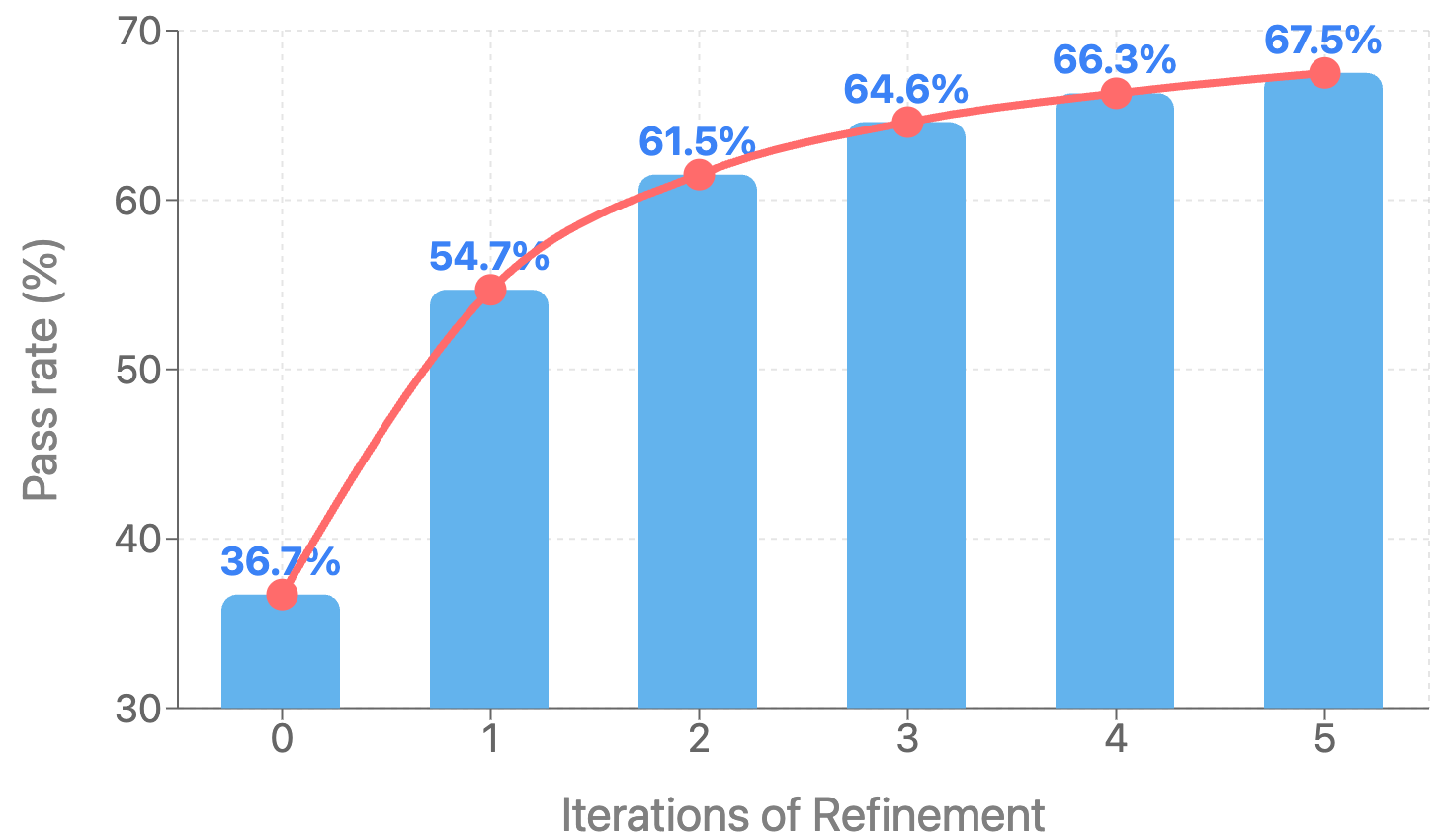}
    \caption{The pass rate progression across iterations of refinement with execution verification feedback.}
    \label{fig:dataset_pass_rate}
\end{figure}

The core of our data generation relies on an iterative refinement process. Figure \ref{fig:dataset_pass_rate} illustrates the progression of code sample pass rates through successive refinement iterations using execution verification feedback.

As depicted in Figure \ref{fig:dataset_pass_rate}, at the first attempt (iter0), only 36.7\% of the code samples pass their corresponding test file, indicating that debugging is necessary for the remaining code. Failed codes go through continual refinement, with the pass rate gradually approaching 67.5\%. The pass rate rises sharply from iter 1 to iter 3 and then slows. Between iter 4 and iter 5, only 1.2\% of cases improve, indicating incremental benefits from further iterations. Thus, additional iterations are not considered.

Current methods construct preference pairs based on pass rate signals and conduct DPO to optimize Code LLMs. Two notable limitations arise: one from the data and one from the algorithm.

Regarding preference data, a snippet with a low pass rate may only require minor modifications to become correct, as errors tend to be isolated to specific parts of the code. We address this by iterative debugging with editing traces naturally annotated by the differences between iterations.

Regarding the algorithm, relying solely on full preference learning can introduce noise during optimization, as positive and negative pairs can be highly similar. This not only hinders the model from learning more effective error correction patterns but also increases the risk of overfitting. We solve this by explicitly identifying which parts of the code need to be aligned.

Iterative debugging can pass through 67.5\% of tasks within a 5-time API call budget for each task. But given just 5 sampling attempts, the pass rate falls to 51.90\% averaged across 5k samples, as shown in Table \ref{tab:debug_sample_pass_rate}. This initial comparison highlights that iterative debugging can achieve a higher pass rate under similar API constraints.
\begin{table}[!htbp]
\centering
\small
\begin{tabular}{llc}
\toprule
 & \textbf{API Calls} & \textbf{Pass rate} (\%) \\
\midrule
Debugging (Ours) & Up to 5 times & \textbf{67.5} \\
Sampling & 5 times for each task & 51.9 \\
\bottomrule
\end{tabular}
\caption{Pass Rate of Debugging and Sampling for Preference Pair Construction.}
\label{tab:debug_sample_pass_rate}
\end{table}

To further investigate the limits of sampling for difficult cases, we collected 5k samples that didn't succeed within 5 debugging iterations. Table \ref{tab:sampling_pass_rate_difficult} details the pass rate when applying N sampling solutions to these difficult tasks.
\begin{table}[!b]
\centering
\small
\begin{tabular}{lcc}
\toprule
\textbf{Sampled Solutions N} & \textbf{Passed} & \textbf{Failed} \\
\midrule
5 & 2.42\% & 97.58\% \\
10 & 3.26\% & 96.26\% \\
15 & 3.96\% & 96.04\% \\
30 & 4.20\% & 95.80\% \\
50 & 4.36\% & 95.64\% \\
\bottomrule
\end{tabular}
\caption{Pass Rate of Sampling for Difficult Cases (Failed within 5 Debugging Iterations).}
\label{tab:sampling_pass_rate_difficult}
\end{table}

The results in Table \ref{tab:sampling_pass_rate_difficult} indicate that for samples which could not succeed with 5 iterations using interpreter feedback and runtime error information, additional sampling alone yields very low pass rates (e.g., only 3.96\% pass with 15 sampling attempts). This suggests such cases can hardly pass through additional sampling alone.

To directly compare the effectiveness of preference pairs generated via iterative debugging versus sampling, we conducted experiments using 10k training samples. Table \ref{tab:results_sampling_debugging_10k} presents these comparative results.
\begin{table*}[!htbp]
\centering
\small
\begin{tabular}{lcccccccc}
\toprule
 & \textbf{HumanEval} & \textbf{HumanEval+} & \textbf{MBPP} & \textbf{MBPP+} & \textbf{BCB-Inst} & \textbf{BCB-Comp} & \textbf{LCB-v5} & \textbf{Avg.} \\
\midrule
Ref. & 61.6 & 53.0 & 76.9 & 62.9 & 40.2 & 45.8 & 24.1 & 52.1 \\
Sampling & 84.1 & 79.3 & 82.3 & 68.5 & 40.8 & 49.2 & 32.4 & 62.3 \\
Debugging & \textbf{87.8} & \textbf{84.8} & \textbf{83.3} & \textbf{69.7} & \textbf{41.1} & \textbf{49.6} & \textbf{32.5} & \textbf{64.1} \\
\bottomrule
\end{tabular}
\caption{Results of Preference Pair Construction using Sampling vs. Debugging (10k training samples).}
\label{tab:results_sampling_debugging_10k}
\end{table*}
As shown in Table \ref{tab:results_sampling_debugging_10k}, iterative debugging can generate more meaningful preference pairs than sampling by leveraging interpreter feedback and runtime information, thereby achieving better results (e.g., an average score of 64.1 vs 62.3) with lower API costs.

\subsection{Synthetic Test Case Generation}
\label{subsec:appendix_synthetic_test_generation}

\subsubsection{Rationale for using Synthetic Test Cases}
\label{subsubsec:appendix_synthetic_rationale}
We address the relationale behind systhetic test cases from the following perpectives.
Optimizing code LLMs through preference learning requires a large amount of training data, which is difficult to annotate or verify manually. Synthetic data has become a widely adopted approach. For example, Qwen2.5-Coder utilizes tens of millions of synthetic instruction samples, and models like DeepSeek-V3 and R1 also incorporate synthetic data during training, also as demonstrated in studies like PLUM, SelfCodeAlign and DSTC.

\subsubsection{Validity of Synthetic Test Cases}
\label{subsubsec:appendix_synthetic_validity}
We have made the following efforts to ensure the qaulity of test cases:
(i) First, we adopted a powerful LLM, GPT-4o, as the test case generator to primarily ensure its validity.
(ii) Through prompting engineering, we have invested significant effort into making the generated test cases broad and meaningful.
(iii) We conducted a manual evaluation by performing a random sample check. We manually examined 100 data samples and found that all the generated test cases correctly reflected the task requirements. However, we observed that these test cases tend to be relatively simple and may not cover all edge cases.

\subsubsection{Coverage Analysis of Test Cases}
\label{subsubsec:appendix_synthetic_coverage}
To validate the effectiveness of test cases in exercising source code, we conducted coverage analysis on a sample of 1,000 training instances. Code coverage, a crucial metric in software testing, quantifies the extent to which a program's source code is exercised by test cases. This metric measures the percentage of code executed by a test suite, which we evaluated using the Python Coverage library\footnote{\url{https://github.com/nedbat/coveragepy}}.
Our evaluation framework organizes the source code and test cases into directories, maintaining a clear separation between the source code (implemented as Python modules) and the corresponding unit tests (developed using the unittest\footnote{\url{https://docs.python.org/3/library/unittest.html}} framework). The coverage metric is calculated through the following equation:

{\small
\begin{equation}
\begin{split}
&\text{Code Coverage} = 100\%  \times \\ 
&\left( \frac{\text{Number of lines of code executed}}{\text{Total Number of lines of code in system component}} \right).
\end{split}
\end{equation}
}

\begin{figure}
    \centering
    \includegraphics[width=1\linewidth]{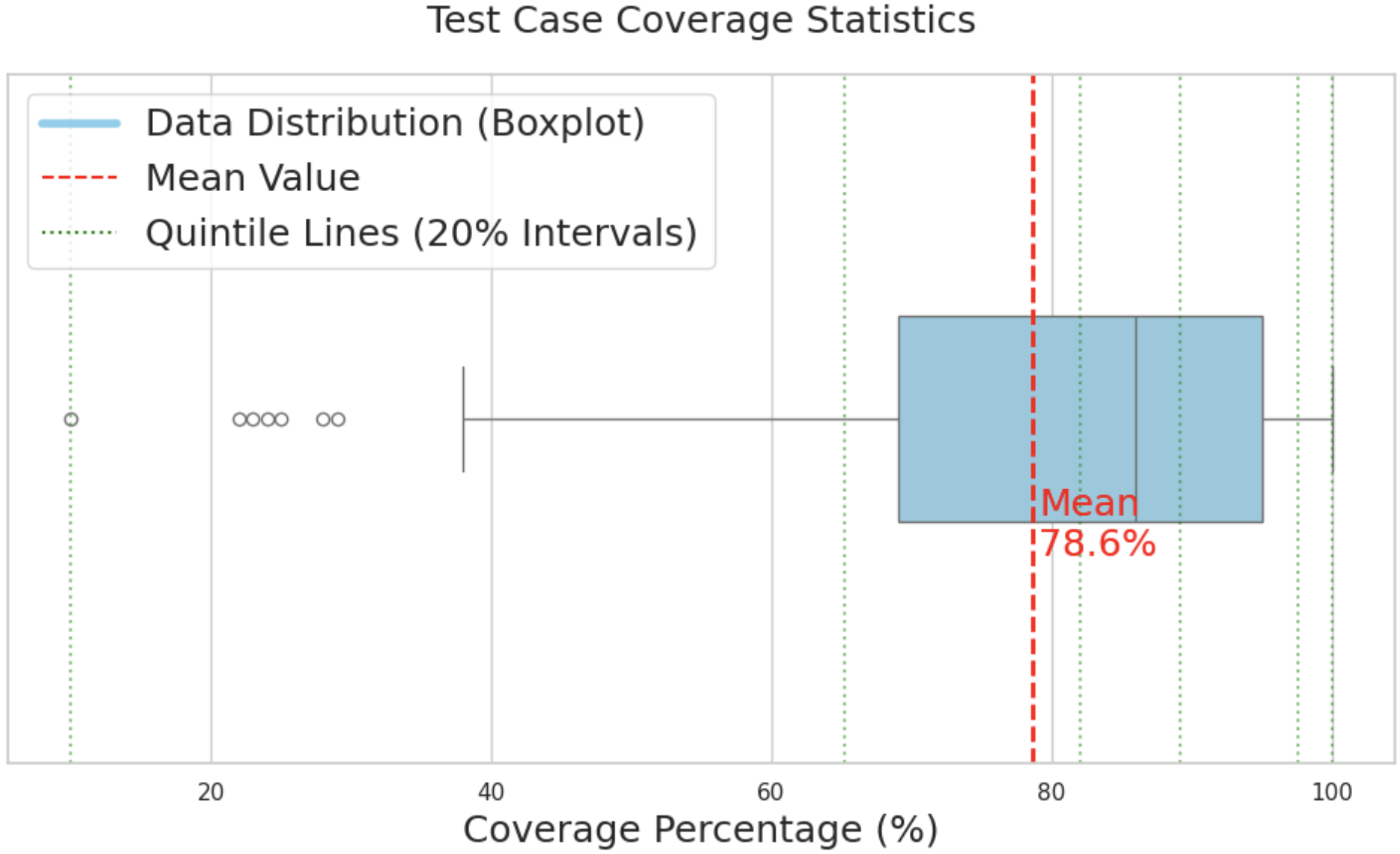}
    \caption{Coverage Distribution of Test Case.}
    \label{fig:coverage}
\end{figure}

The results of the coverage analysis are shown in Figure \ref{fig:coverage}.

\subsubsection{Evaluating the quality of test cases using LLM-as-a-judge}
\begin{figure}[h]
    \centering
    \includegraphics[width=0.98\linewidth]{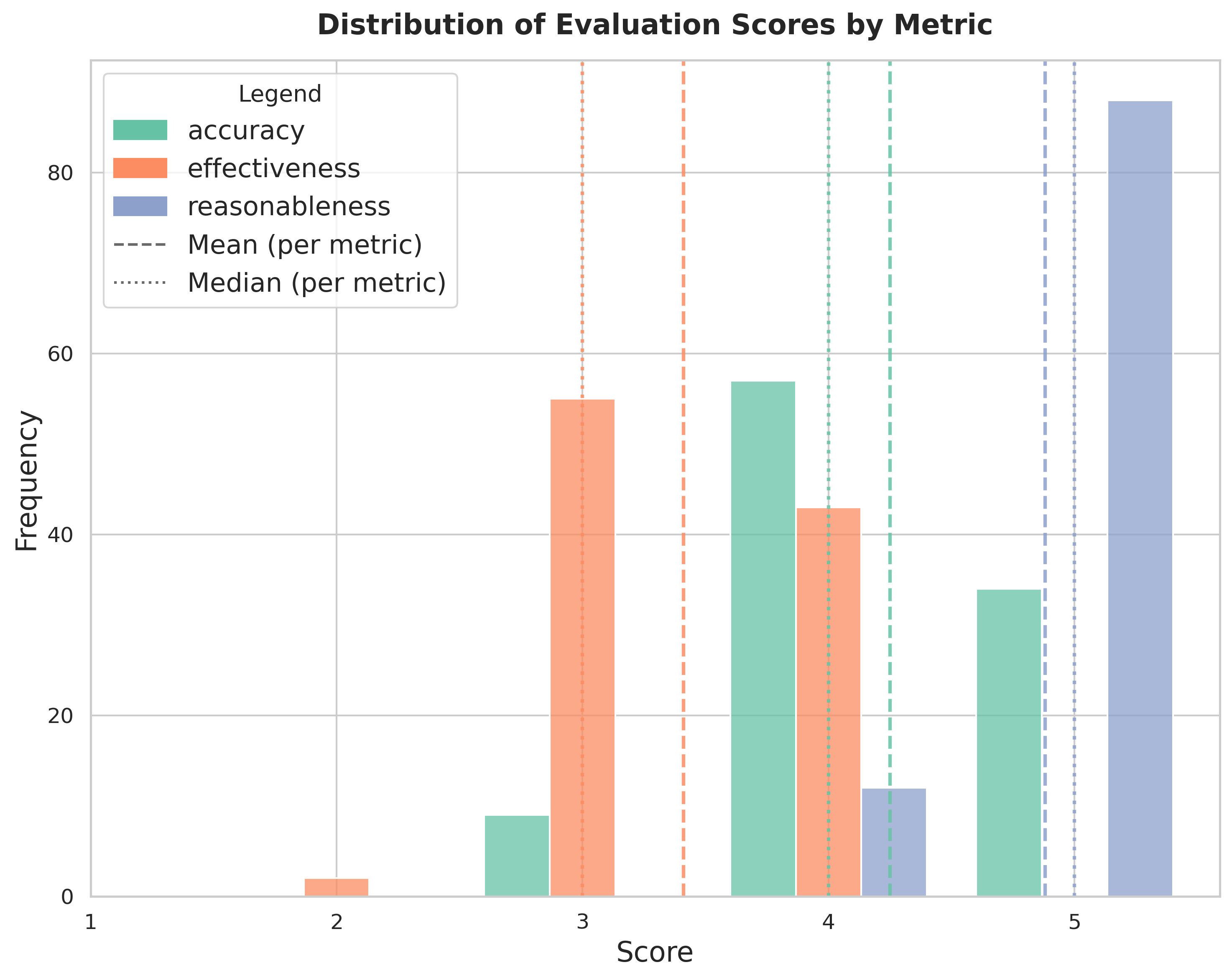}
    \caption{Comparative distribution of LLM-as-a-judge scores across three evaluation metrics: accuracy, effectiveness, and reasonableness.}
    \label{fig:dist_llmasjudge}
\end{figure}

\begin{figure*}
    \centering
    \includegraphics[width=1\linewidth]{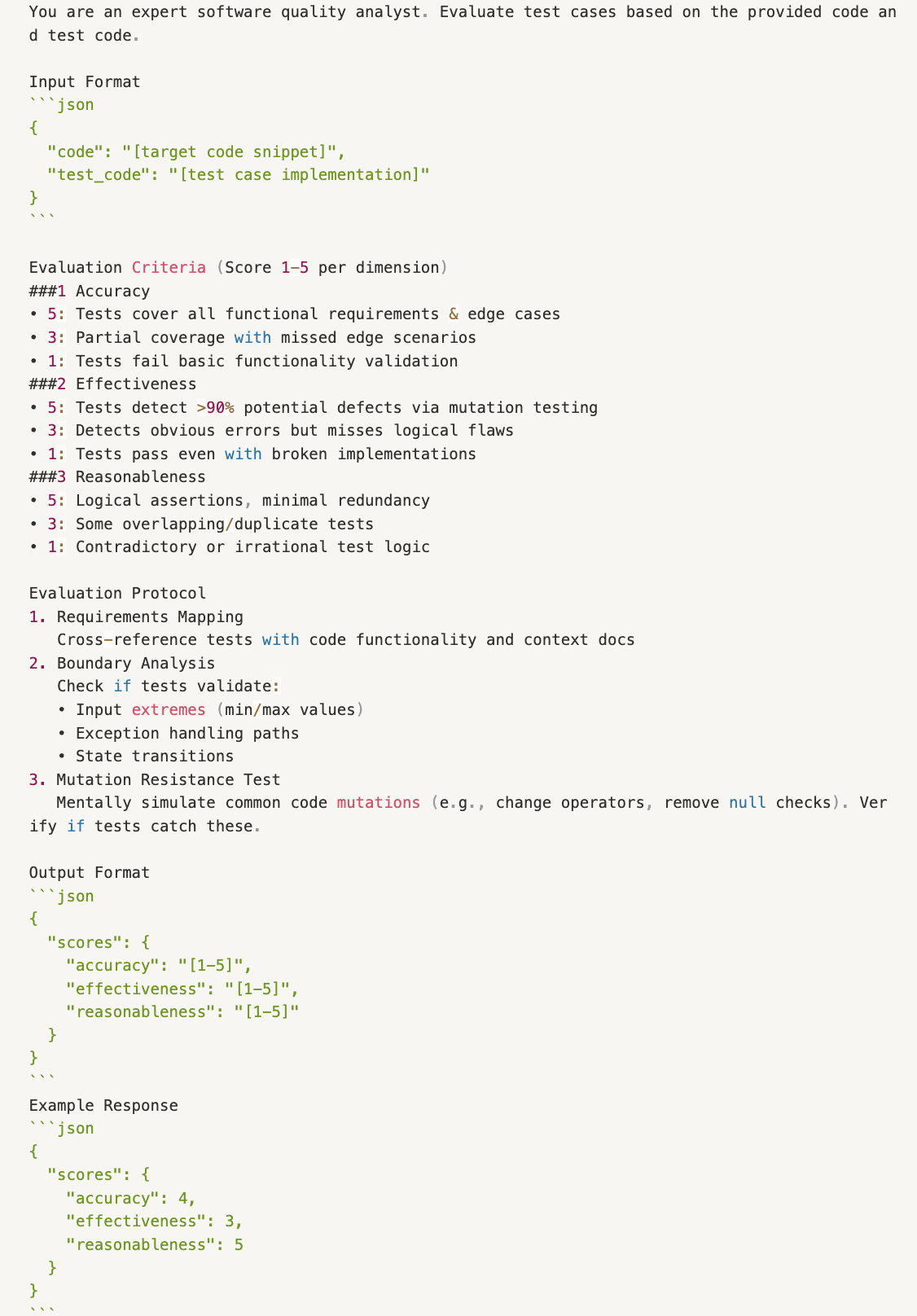}
    \caption{Prompt for evaluating the quality of test cases using LLM-as-a-judge.}
    \label{fig:prompt_llmasjudge}
\end{figure*}

To evaluate test case quality, we employ the LLM-as-a-judge approach, assessing three dimensions: \textbf{accuracy}, \textbf{effectiveness}, and \textbf{reasonableness} on a 5-point scale. Detailed evaluation prompts are provided in \ref{fig:prompt_llmasjudge}. We sample 1,000 data points from the training data and conduct evaluations using the \texttt{DeepSeek-V3-0324} model. The evaluation results in Figure~\ref{fig:dist_llmasjudge} demonstrate satisfactory test case quality across all dimensions.

\subsubsection{Prompt and Examples for Generated Test Cases}
\label{subsubsec:appendix_synthetic_prompt_examples}
\begin{figure*}[!htbp]
    \centering
    \includegraphics[width=1\linewidth]{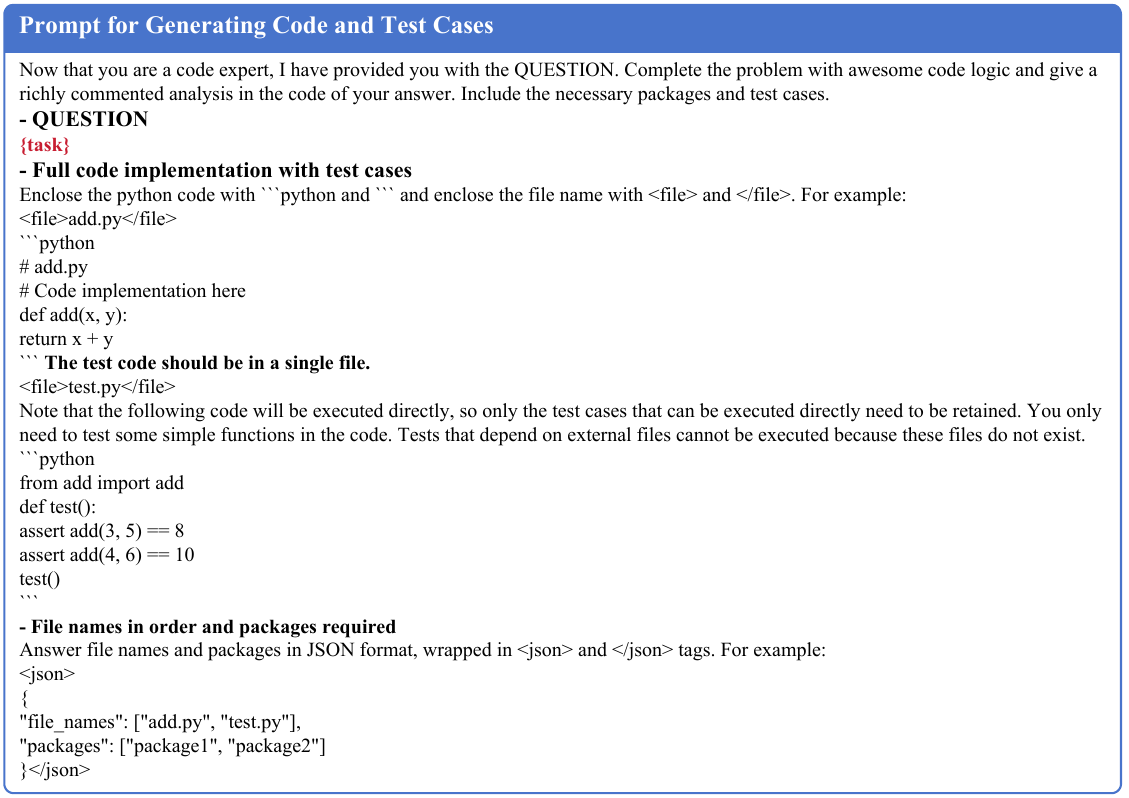}
    \caption{Prompt used for generating code and test cases.}
    \label{fig:testcase_prompt}
\end{figure*}

\begin{figure*}[!htbp]
    \centering
    \includegraphics[width=0.84\linewidth]{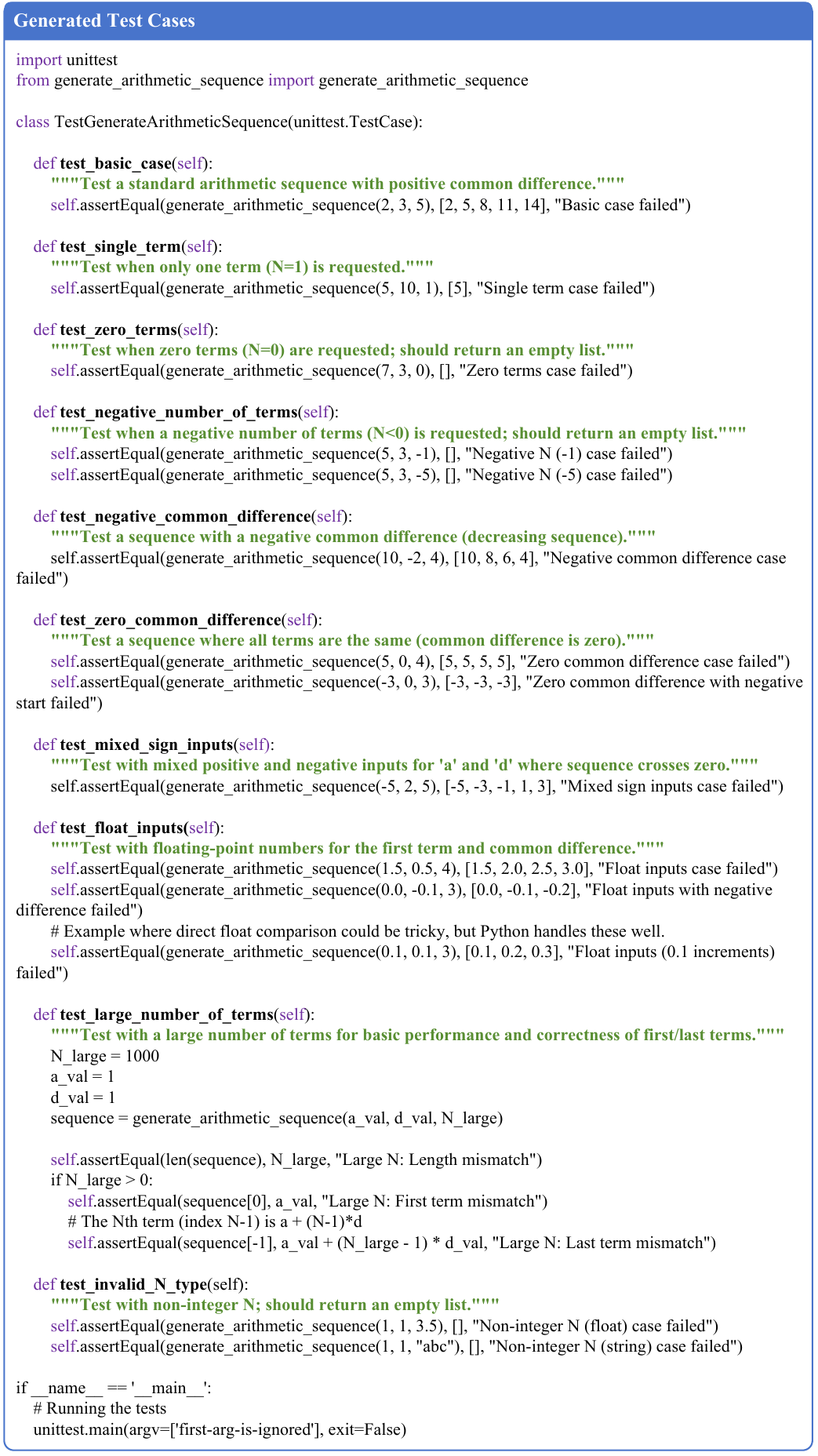}
    \caption{An example of generated test cases.}
    \label{fig:generated_testcases_example}
\end{figure*}

To illustrate our synthetic test case generation process, Figure \ref{fig:testcase_prompt} displays the prompt template provided to the LLM. Following this prompt, Figure \ref{fig:generated_testcases_example} presents an example of the test cases generated for question ``Create a Python function named `generate\_arithmetic\_sequence' that generates the first N terms of an arithmetic sequence given the first term and the common difference. The function should take three parameters: the first term (a), the common difference (d), and the number of terms (N). The function should return a list containing the first N terms of the sequence”.

\subsection{Training and Inference Parameters}
\label{subsec:appendix_training_inference_params}

Unless specified otherwise for a particular experiment, our code LLMs were trained with consistent hyperparameter settings. The learning rate was set to $1 \times 10^{-5}$ for the 7B code LLMs and $5 \times 10^{-6}$ for the 14B models. We used a global batch size of 128, full-parameter training 3 epochs with max prompt length of 1024 and generation length of 2048. A cosine learning rate scheduler with was employed, with 3\% of the total training steps dedicated to warm-up. For the DPO algorithm, $\beta$ is set to 0.1, and $\alpha$ is set to 1.0 for RPO. $\pi_\theta$ and $\pi_{\text{ref}}$ are both initialized with the weights of the evaluated model, while $\pi_{\text{ref}}$ keeps frozen during training.
For all inference, greedy decoding was utilized with pass rate at first attempt reported. For Code-Optimise, we replicated their setup, and the $DPO_{PvF}$ setting results are reported. For PLUM and CodeDPO, their scores are taken from their respective papers.


\section{Experimental Results and Comparisons}
\label{sec:appendix_experimental_results_comparisons}

This section begins by describing the evaluation benchmarks and their statistics. It then presents the comprehensive experimental results of our method on these benchmarks, followed by detailed comparisons against Supervised Fine-Tuning (SFT) and other relevant state-of-the-art methods.


\subsection{Evaluation Benchmarks: Description and Statistics}
\label{subsec:appendix_evaluation_benchmarks}
We detail the individual function-level code generation benchmarks used for evaluation in this subsection. Table \ref{tab:benchmark_stats} summarizes key statistics for these benchmarks, such as the number of problems and the average number of tests per problem.
\begin{table}[!htbp]
\centering
\small
\begin{tabular}{llc}
\toprule
\textbf{Dataset} & \textbf{Problems} & \textbf{Avg. Tests} \\
\midrule
HumanEval & \multirow{2}{*}{164} & 9.57 \\
HumanEval+ & & 748.07 \\
\cmidrule(lr){2-3}
MBPP & \multirow{2}{*}{378} & 3.11 \\
MBPP+ & & 105.40 \\
\cmidrule(lr){2-3}
 & Easy~~~279 & 18.07 \\
LiveCodeBench & Medium~331 & 21.81 \\
 & Hard~~~270 & 24.78 \\
\bottomrule
\end{tabular}
\caption{Statistics of Evaluation Benchmarks.}
\label{tab:benchmark_stats}
\end{table}

\textbf{HumanEval and MBPP} are popular benchmarks for assessing code generation. Considering the limited test cases in these benchmarks (HumanEval: 9.57 avg. tests; MBPP: 3.11 avg. tests, as seen in Table \ref{tab:benchmark_stats}), we followed previous work and utilized the EvalPlus framework to evaluate model robustness across a broader range of test cases (HumanEval+: 748.07 avg. tests; MBPP+: 105.40 avg. tests). To ensure fair comparison, we used version 0.2.0 of MBPP+ provided by EvalPlus\footnote{\url{https://github.com/evalplus/evalplus}} v0.3.1 , which removes some broken tasks (399 $\to$ 378 tasks).

\textbf{BigCodeBench (BCB)} is a comprehensive benchmark designed to assess a model’s ability to handle real-world programming tasks, particularly its effectiveness in utilizing various function calls as tools. Our model's ability to adeptly manage these high-complexity scenarios underscores its suitability for BigCodeBench.

\textbf{LiveCodeBench (LCB)}, statistics for which are also included in Table \ref{tab:benchmark_stats} (showing problem distribution by difficulty), is a benchmark designed to evaluate code generation models on challenging competitive programming problems, often sourced from real coding contests. Unlike benchmarks focused solely on function completion, LCB tasks typically require more complex algorithmic reasoning and problem-solving skills. The evaluation often simulates a contest environment, potentially including hidden test cases to assess the robustness and correctness of the generated solutions under pressure. In our experiments, we utilize LiveCodeBench-v5 (LCB-v5) to gauge the model's capabilities in tackling these demanding, contest-style coding scenarios.

\subsection{Comparison with Supervised Fine-Tuning}
\label{subsec:appendix_comparison_sft}
We compare our method with standard Supervised Fine-Tuning (SFT) to demonstrate the benefits of our preference alignment approach. The overall comparison of our method (Target-RPO) against SFT and reference models on various benchmarks is presented in Table \ref{tab:sft_comparison_main}. For instance, on Qwen2.5-Coder-7B, our Target-RPO achieves an average of 65.3, surpassing the SFT baseline's 63.2.
\begin{table*}[!htbp]
\centering
\small
\begin{tabular}{lcccccccc}
\toprule
\multirow{2}{*}{} & \multicolumn{2}{c}{\textbf{HumanEval}} & \multicolumn{2}{c}{\textbf{MBPP}} & \multirow{2}{*}{\textbf{BCB-Inst}} & \multirow{2}{*}{\textbf{BCB-Comp}} & \multirow{2}{*}{\textbf{LCB-v5}} & \multirow{2}{*}{\textbf{Average}} \\
\cmidrule(lr){2-3} \cmidrule(lr){4-5}
 & \textit{\textbf{Base}} & \textit{\textbf{Plus}} & \textit{\textbf{Base}} & \textit{\textbf{Plus}} & & & & \\
\midrule
\multicolumn{9}{c}{\textbf{\textit{CodeQwen1.5-7B-Chat}}} \\
Ref. & 83.5 & 78.7 & 79.4 & 69.0 & 39.6 & 43.6 & 15.3 & 58.4 \\
SFT & 87.8 & 83.5 & 82.3 & 69.6 & 35.9 & 45.6 & 17.0 & 60.2 \\
Target-DPO & 89.6 & 85.4 & \textbf{83.9} & 69.8 & \textbf{39.9} & \textbf{48.7} & \textbf{20.2} & \textbf{62.5} \\
Target-RPO & \textbf{89.6} & \textbf{86.0} & 82.5 & \textbf{70.4} & 38.3 & 48.4 & 19.9 & 62.2 \\
\midrule
\multicolumn{9}{c}{\textit{\textbf{Qwen2.5-Coder-7B}}} \\
Ref. & 61.6 & 53.0 & 76.9 & 62.9 & 40.2 & 45.8 & 24.1 & 52.1 \\
SFT & 87.2 & 82.9 & 83.1 & 68.3 & 39.1 & 51.6 & 30.0 & 63.2 \\
DiffAug-RPO & 86.0 & 81.7 & 82.8 & 67.5 & 40.7 & 51.4 & 30.5 & 62.9 \\
Target-RPO  & \textbf{89.6} & \textbf{84.8} & \textbf{83.3} & \textbf{69.5} & \textbf{43.1} & \textbf{53.3} & \textbf{33.3} & \textbf{65.3} \\
\bottomrule
\end{tabular}
\caption{SFT and our results on CodeQwen1.5-7B-Chat and Qwen2.5-Coder-7B. Detailed results on LiveCodeBench and BigCodeBench are presented in Table~\ref{tab:sft_lcb_detailed} and Table~\ref{tab:sft_bcb_detailed}.}
\label{tab:sft_comparison_main}
\end{table*}

Table \ref{tab:sft_lcb_detailed} provides a more granular breakdown of performance on LiveCodeBench-v5 by difficulty, where our method consistently outperforms the SFT versions of both CodeQwen-7B-Chat and Qwen-Coder-7B, especially on Easy and Medium problems.
\begin{table*}[!htbp]
\centering
\small
\begin{tabular}{lcccc}
\toprule
 & \textbf{EASY} & \textbf{MEDIUM} & \textbf{HARD} & \textbf{Avg} \\
\midrule
CodeQwen-7B-Chat-SFT & 41.87 & 9.14 & 0.75 & 17.06 \\
CodeQwen-7B-Chat-Ours & \textbf{48.73} & \textbf{11.89} & \textbf{0.75} & \textbf{20.16} \\
Qwen-Coder-7B-SFT & 67.14 & 21.03 & 2.61 & 30.01 \\
Qwen-Coder-7B-Ours & \textbf{69.31} & \textbf{27.13} & \textbf{3.73} & \textbf{33.33} \\
\bottomrule
\end{tabular}
\caption{Detailed Results on LiveCodeBench-v5, comparing SFT with Our \method.}
\label{tab:sft_lcb_detailed}
\end{table*}

Similarly, Table \ref{tab:sft_bcb_detailed} shows detailed results on BigCodeBench categories, again illustrating the advantage of our approach over SFT across different task types and difficulties. For example, Qwen-Coder-7B with our method achieves an average of 36.8 compared to SFT's 31.8.
\begin{table*}[!htbp]
\centering
\small
\begin{tabular}{lccccc}
\toprule
 & \textbf{Complete-Full} & \textbf{Instruct-Full} & \textbf{Complete-Hard} & \textbf{Instruct-Hard} & \textbf{Average} \\
\midrule
CodeQwen-7B-Chat-SFT & 45.6 & 35.9 & 18.3 & 15.5 & 28.8 \\
CodeQwen-7B-Chat-Ours & \textbf{48.4} & \textbf{38.3} & \textbf{20.3} & \textbf{18.2} & \textbf{31.3} \\
Qwen-Coder-7B-SFT & 51.6 & 39.1 & 21.6 & 14.9 & 31.8 \\
Qwen-Coder-7B-Ours & \textbf{53.3} & \textbf{43.1} & \textbf{29.7} & \textbf{20.9} & \textbf{36.8} \\
\bottomrule
\end{tabular}
\caption{Detailed Results on BigCodeBench, comparing SFT with Our \method.}
\label{tab:sft_bcb_detailed}
\end{table*}

\subsection{Comparison with Other Code Generation Methods}
\label{subsec:appendix_comparison_other_methods}
We further benchmark our method against other notable code generation techniques. Table \ref{tab:code-performance} benchmarks our method against techniques like CodeDPO, Code-Optimise, and PLUM on HumanEval and MBPP. Notably, on the DeepSeekCoder-6.7B base, our method achieves significantly higher scores (e.g., 66.50 on HumanEval vs. 59.75 for CodeDPO and 56.70 for PLUM).

\begin{table*}[!htbp]
\centering
\small
\begin{tabular}{lcccc}
\toprule
 & \textbf{HumanEval} & \textbf{HumanEval+} & \textbf{MBPP} & \textbf{MBPP+} \\
\midrule
StarCoder2-7B & 35.40 & 29.90 & 54.40 & 45.60 \\
CodeDPO & 48.17 & 34.15 & 58.40 & 49.37 \\
Code-Optimise & 32.32 & 28.05 & 58.90 & 47.89 \\
PLUM & 46.30 & 39.60 &  60.40 & 49.10 \\
Our \method~ & \textbf{48.20} & \textbf{43.90} & \textbf{63.50} & \textbf{50.60} \\
\midrule
DeepSeekCoder-1.3B & 31.53 & 28.65 & 57.40 & 48.67 \\
CodeDPO & 42.07 & 38.04 & 61.37 & 53.43 \\
Code-Optimise & 34.15 & 30.49 & 59.15 & 49.87 \\
Our \method~ & \textbf{47.00} & \textbf{43.30} & \textbf{61.37} & \textbf{54.20}  \\
\midrule
DeepSeekCoder-6.7B & 47.60 & 39.60 & 70.20 & 56.60 \\
CodeDPO & 59.75 & 51.83 & 72.18 & 60.01 \\
Code-Optimise & 47.56 & 37.20 & 72.18 & 57.64 \\
PLUM & 56.70 & 48.80 & 72.90 & 58.90 \\
Our \method & \textbf{66.50} & \textbf{60.40} & \textbf{76.50} & \textbf{61.40} \\
\bottomrule
\end{tabular}
\caption{Performance comparison with baselines.}
\label{tab:code-performance}
\end{table*}



\section{In-depth Analyses and Ablation Studies}
\label{sec:appendix_in_depth_ablations}

In this section, we conduct several in-depth analyses and ablation studies to better understand the characteristics and behavior of our proposed method. This includes investigating the impact of model size (scaling laws), sensitivity to key hyperparameters ($\beta$ and $\alpha$), the diversity of our constructed preference data, common error patterns in the generated code, and the efficiency of our preference annotation process.

\subsection{Scaling Law on Model Size}
\label{subsec:appendix_scaling_model_size_results}
The impact of model size on performance when applying our method is detailed in Table \ref{tab:size-ablation}. The results show a clear trend: as model size increases from 1.5B to 32B parameters, the average performance improves from 54.4 to 74.7, demonstrating the scalability of our approach.
\begin{table*}[!htbp]
\centering
\small
\begin{tabular}{lccccccc}
\toprule
\multirow{2}{*}{Model} & \multicolumn{2}{c}{\textbf{HumanEval}} & \multicolumn{2}{c}{\textbf{MBPP}} & \multirow{2}{*}{\textbf{BCB-Inst}} & \multirow{2}{*}{\textbf{BCB-Comp}} & \multirow{2}{*}{\textbf{Average}} \\
\cmidrule(lr){2-3} \cmidrule(lr){4-5}
 & \textit{\textbf{Base}} & \textit{\textbf{Plus}} & \textit{\textbf{Base}} & \textit{\textbf{Plus}} & & & \\
\midrule
Qwen2.5-Coder-1.5B & 67.7 & 62.8 & 66.7 & 55.3 & 33.4 & 40.5 & 54.4 \\
Qwen2.5-Coder-7B  & 89.6 & 84.8 & 83.3 & 69.5 & 43.1 & 53.3 &  70.6 \\
Qwen2.5-Coder-32B & \textbf{92.7} & \textbf{86.6} & \textbf{89.4} & \textbf{74.6} & \textbf{45.7} & \textbf{58.9} & \textbf{74.7} \\
\bottomrule
\end{tabular}
\caption{Ablations on model size.}
\label{tab:size-ablation}
\end{table*}

\subsection{Ablation Studies on Hyperparameters \texorpdfstring{$\beta$ and $\alpha$}{beta and alpha}}
\label{subsec:appendix_ablations_beta_alpha_results}
In Direct Preference Optimization (DPO), the hyperparameter $\beta$ controls the strength of the preference signal, essentially determining how strictly the model should adhere to the learned preferences relative to the reference model. The hyperparameter $\alpha$, when part of the DPO framework or a combined loss, often serves as a weighting factor for an additional objective or regularization term. 
We performed ablation studies on key hyperparameters $\beta$ and $\alpha$. Table \ref{tab:beta-ablation} presents the results for $\beta$ when $\alpha$ is set to 0, suggesting that a smaller $\beta$ (e.g., 0.1) yields the best average performance (69.7).
\begin{table*}[!htbp]
\centering
\small
\begin{tabular}{lccccccc}
\toprule
\multirow{2}{*}{$\beta$} & \multicolumn{2}{c}{\textbf{HumanEval}} & \multicolumn{2}{c}{\textbf{MBPP}} & \multirow{2}{*}{\textbf{BCB-Inst}} & \multirow{2}{*}{\textbf{BCB-Comp}} & \multirow{2}{*}{\textbf{Average}} \\
\cmidrule(lr){2-3} \cmidrule(lr){4-5}
 & \textit{\textbf{Base}} & \textit{\textbf{Plus}} & \textit{\textbf{Base}} & \textit{\textbf{Plus}} & & & \\
\midrule
0.1 & \textbf{89.0} & \textbf{83.6} & 83.1 & 69.0 & \textbf{41.0} & \textbf{52.7} & \textbf{69.7} \\
0.3 & 86.6 & 80.5 & 82.5 & \textbf{68.5} & 40.3 & 50.3 & 69.1 \\
0.5 & 85.4 & 80.5 & \textbf{83.6} & 66.7 & 40.5 & 48.2 & 67.5 \\
\bottomrule
\end{tabular}
\caption{Ablations on $\beta$ with $\alpha$ set 0.}
\label{tab:beta-ablation}
\end{table*}

The corresponding ablation for $\alpha$, with $\beta$ fixed at 0.1, is shown in Table \ref{tab:alpha_ablation_beta_0_1}. These results indicate that $\alpha=1.0$ provides the highest average score (70.5), while $\alpha=\infty$ (equivalent to SFT) performs relatively worse.
\begin{table*}[!htbp]
\centering
\small
\begin{tabular}{lccccccc}
\toprule
\multirow{2}{*}{$\alpha$} & \multicolumn{2}{c}{\textbf{HumanEval}} & \multicolumn{2}{c}{\textbf{MBPP}} & \multirow{2}{*}{\textbf{BCB-Inst}} & \multirow{2}{*}{\textbf{BCB-Comp}} & \multirow{2}{*}{\textbf{Average}} \\
\cmidrule(lr){2-3} \cmidrule(lr){4-5}
 & \textit{\textbf{Base}} & \textit{\textbf{Plus}} & \textit{\textbf{Base}} & \textit{\textbf{Plus}} & & & \\
\midrule
1.0 & \textbf{89.6} & \textbf{84.8} &  \textbf{83.1} &  69.3 & \textbf{43.1} & 53.3 & \textbf{70.5} \\
3.0 & 87.2 & 81.1 & 82.8 & 69.0 & 40.3 & 52.7 & 67.9 \\
5.0 & 84.1 & 80.5 & 83.1 & \textbf{70.1} & 40.3 &  \textbf{54.0} & 68.7 \\
$\infty$ (SFT) & 87.2 & 82.9 & 83.1 & 68.3 & 39.1 & 51.6  &  68.7 \\
\bottomrule
\end{tabular}
\caption{Ablations on $\alpha$ with $\beta$ set 0.1.}
\label{tab:alpha_ablation_beta_0_1}
\end{table*}

\subsection{Ablation on Context Usage}
\label{subsec:appendix_qwen_ablation_results}

Beyond CodeQwen1.5-7B-Chat, we also provide ablation results on the use of context from the rejected sample using Qwen2.5-Coder-7B, as shown in Table~\ref{ablation_qwen}.

\begin{table*}[!t]
\centering
\small
\begin{tabular}{lcccccccc}
\toprule
\multirow{2}{*}{} & \multicolumn{2}{c}{\textbf{HumanEval}} & \multicolumn{2}{c}{\textbf{MBPP}} & \multirow{2}{*}{\textbf{BCB-Inst}} & \multirow{2}{*}{\textbf{BCB-Comp}} & \multirow{2}{*}{\textbf{LCB-v5}} & \multirow{2}{*}{\textbf{Average}} \\
\cmidrule(lr){2-3} \cmidrule(lr){4-5}
 & \textit{\textbf{Base}} & \textit{\textbf{Plus}} & \textit{\textbf{Base}} & \textit{\textbf{Plus}} & & & & \\
\midrule
Qwen2.5-Coder-7B & 61.6 & 53.0 & 76.9 & 62.9 & 40.2 & 45.8 & 24.1 & 52.1 \\
SFT & 87.2 & 82.9 & 83.1 & 68.3 & 39.1 & 51.6 & 30.0 & 63.2 \\
Hybrid-RPO & 82.9 & 79.3 & 81.7 & 67.5 & 41.2 & 50.5 & 29.8 & 61.8 \\
DiffAug-RPO & 86.0 & 81.7 & 82.8 & 67.5 & 40.7 & 51.4 & 30.5 & 62.9 \\
Target-RPO & \textbf{89.6} & \textbf{84.8} & \textbf{83.3} & \textbf{69.5} & \textbf{43.1} & \textbf{53.3} & \textbf{33.3} & \textbf{65.3} \\
\bottomrule
\end{tabular}
\caption{Ablation results on the \method~using Qwen2.5-Coder-7B.}
\label{ablation_qwen}
\end{table*}

\subsection{Analysis of Preference Data Diversity}
\label{subsec:appendix_data_diversity_results}
To understand the characteristics of our CodeFlow preference dataset, Table \ref{tab:feature_distribution} provides a distributional analysis of various features (e.g., Workflow, Functionality, Data Processing) across 1k samples, comparing it with other common datasets like Alpaca and OSS-Instruct. Our CodeFlow dataset (both preferred and dis-preferred samples) generally exhibits a higher count and thus potentially greater diversity across most features, particularly in Data Processing, File Operation, and Advanced Techniques.
\begin{table*}[!htbp]
\centering
\small
\begin{tabular}{lcccccc}
\toprule
\textbf{Datasets} & \textbf{Workflow} & \textbf{Functionality} & \begin{tabular}[c]{@{}c@{}}\textbf{Computation}\\\textbf{Operation}\end{tabular} & \begin{tabular}[c]{@{}c@{}}\textbf{User}\\\textbf{Interaction}\end{tabular} & \begin{tabular}[c]{@{}c@{}}\textbf{Data}\\\textbf{Processing}\end{tabular} & \begin{tabular}[c]{@{}c@{}}\textbf{File}\\\textbf{Operation}\end{tabular} \\
\midrule
Alpaca & 994 & 393 & 282 & 82 & 221 & 11 \\
CodeFeedback & 2079 & 535 & 689 & 143 & 895 & 39 \\
Evol-Alpaca & 2163 & 591 & 783 & 134 & 1401 & 55 \\
OSS-Instruct & 2254 & 669 & 413 & 192 & 903 & 102 \\
CodeFlow (Preferred)  & \textbf{2689} & \textbf{805} & \textbf{967} & \textbf{410} & \textbf{2418} & \textbf{290} \\
CodeFlow (Dis-Preferred)  & 2490 & 772 & 964 & 406 & 2327 & 287 \\
\bottomrule
\end{tabular}

\vspace{0.5em}

\begin{tabular}{lccccccc}
\textbf{} & \textbf{Logging} & \textbf{Algorithm} & \begin{tabular}[c]{@{}c@{}}\textbf{Data}\\\textbf{Structures}\end{tabular} & \begin{tabular}[c]{@{}c@{}}\textbf{Implementation}\\\textbf{Logic}\end{tabular} & \begin{tabular}[c]{@{}c@{}}\textbf{Advanced}\\\textbf{Techniques}\end{tabular} & \textbf{Average} \\
\midrule
Alpaca & 1 & 232 & 72 & 67 & 10 & 215.00 \\
CodeFeedback & 10 & 427 & 100 & 49 & 63 & 457.18 \\
Evol-Alpaca & 15 & 414 & 130 & 74 & 94 & 532.18 \\
OSS-Instruct & 62 & 150 & 140 & 82 & 26 & 453.91 \\
CodeFlow (Preferred)  & \textbf{133} & \textbf{790} & \textbf{367} & \textbf{152} & \textbf{178} & \textbf{836.27} \\
CodeFlow (Dis-Preferred)  & 129 & 785 & 361 & 149 & 193 & 805.73 \\
\bottomrule
\end{tabular}
\caption{Distribution of total features across 1k samples.}
\label{tab:feature_distribution}
\end{table*}

\subsection{Error Analysis of Generated Code}
\label{subsec:appendix_error_analysis_results}
\paragraph{\method~Generates Fewer Errors.}
In this section, we present a statistical analysis of common failure case types to pinpoint frequent pitfalls in code generation.
By contrasting critical tokens between a corrected version and its preceding iteration explicitly, a Code LLM equipped with \method~makes fewer errors. Table \ref{tab:error_comparison} presents the frequency of common failure types (e.g., AttributeError, KeyError) on the BigCodeBench Complete-Full set. Our Target-RPO method shows a notable reduction in the sum of these errors (308 occurrences) compared to RPO (369) and Code-Optimise-RPO (396) on Qwen2.5-Coder-7B. This suggests that while RPO includes SFT, it still requires targeted learning of critical errors in the dispreferred samples to effectively reduce mistakes.
\begin{table*}[!htbp]
    \centering
    \small
    \begin{tabular}{lccc}
        \toprule
        \textbf{Error Type} & \textbf{RPO} & \textbf{Code-Optimise-RPO} & \textbf{Target-RPO (Ours)} \\
        \midrule
        AttributeError: `X' has no attribute `Y' & 145 & 149 & 127 \\
        KeyError: `X' & 139 & 128 & 100 \\
        NameError: name `X' is not defined & 41 & 58 & 46 \\
        FileNotFoundError: No such file or directory & 44 & 61 & 35 \\
        \textbf{Sum} & 369 & 396 & \textbf{308} \\
        \bottomrule
    \end{tabular}
    \caption{The frequency of the most common failure types on the BigCodeBench Complete-Full set.}
    \label{tab:error_comparison}
\end{table*}

\subsection{Efficiency Analysis of Preference Annotation}
\label{subsec:appendix_efficiency_analysis_results}
Additionally, we compare the costs of generating and annotating preference pairs to guide more efficient preference alignment and reduce errors.

\method~Provides an Efficient Pathway for Preference Annotation. We compare the cost of synthesizing one preference pair between \method~and the sampling techniques adopted by Code-Optimise, primarily considering external LLM calls and execution times.
Given an instruction, Code-Optimise synthesizes $m$ code snippet candidates (where $m$ is often set to 100), using $n$ test cases from the raw dataset, leading to $m \times n$ executions on the CPU.
In contrast, for a single instruction, 
\method~requires up to 7 LLM calls and executions for successful pair generation in most cases. Considering the failure ratio (when code can't pass the generated test cases within the budget), an estimated 10.4 calls are needed for a given instruction on average across the dataset. This is far fewer than the $m$ (e.g., 100) calls for sampling candidates plus subsequent executions often employed by sampling-heavy methods.
Though massive sampling can yield diverse candidates, it is not efficient as most code snippets are discarded. \method~shows that starting with a single code snippet, even if it fails initially, it still holds high potential to form a valuable preference pair for alignment training through iterative refinement.

\begin{table*}[!htbp]
\centering
\small
\begin{tabular}{lccccccc|c}
\toprule
Method & HE & HE+ & MBPP & MBPP+ & BCB-Comp & BCB-Inst & LCB-v5 & Avg. \\
\midrule
Ref. & 61.6 & 53.0 & 76.9 & 62.9 & 45.8 & 40.2 & 24.1 & 52.1 \\
Prefix-Suffix & 86.6 & 79.9 & 81.5 & 68.0 & 47.8 & 39.2 & 30.1 & 61.9 \\
LCS & \textbf{89.0} & \textbf{83.6} & \textbf{83.1} & \textbf{69.0} & \textbf{52.7} & \textbf{41.0} & \textbf{32.6} & \textbf{64.4} \\
\bottomrule
\end{tabular}
\caption{Comparison of Prefix-Suffix and LCS methods for code difference extraction across benchmarks.}
\label{code_difference}
\end{table*}

\begin{table*}[!htbp]
\centering
\small
\begin{tabular}{lccccccc|c}
\toprule
Method & HE & HE+ & MBPP & MBPP+ & BCB-Comp & BCB-Inst & LCB-v5 & Avg. \\
\midrule
Ref. & 61.6 & 53.0 & 76.9 & 62.9 & 45.8 & 40.2 & 24.1 & 52.1 \\
First & 87.8 & 81.7 & 81.0 & 67.5 & 47.2 & 40.4 & 29.7 & 62.2 \\
Last & 88.4 & 82.9 & 82.0 & 68.0 & 51.7 & 41.2 & 31.5 & 63.7 \\
Random (Ours) & \textbf{89.0} & \textbf{83.6} & \textbf{83.1} & \textbf{69.0} & \textbf{52.7} & \textbf{41.0} & \textbf{32.6} & \textbf{64.4} \\
\bottomrule
\end{tabular}
\caption{Comparison of different negative sample selection strategies across benchmarks.}
\label{select_negatives}
\end{table*}

\begin{table*}[!thbp]
\centering
\small
\begin{tabular}{lccccccc|c}
\toprule
Method & HE & HE+ & MBPP & MBPP+ & BCB-Comp & BCB-Inst & LCB-v5 & Avg. \\
\midrule
Ref. & 61.6 & 53.0 & 76.9 & 62.9 & 45.8 & 40.2 & 24.1 & 52.1 \\
DS-v3-SFT & 82.3 & 76.2 & 83.3 & 69.8 & 46.9 & 40.7 & 26.4 & 60.8 \\
DS-v3-Ours & \textbf{84.8} & \textbf{79.3} & \textbf{84.7} & \textbf{72.0} & \textbf{47.5} & \textbf{40.7} & \textbf{27.7} & \textbf{62.4} \\
\bottomrule
\end{tabular}
\caption{Results of distillation from DeepSeek-V3 to Qwen2.5-Coder-7B.}
\label{teacher_ablation}
\end{table*}

\subsection{Alternatives to LCS for Code Differencing}

To explore alternatives beyond LCS-based code differencing, we considered a simple baseline approach where we extract the common prefix and suffix, treating the middle part as the difference. The results are shown in Table~\ref{code_difference}:

Consider these two versions of a Python function where only two lines are truly different:

\begin{lstlisting}[language=Python, caption={Chosen Code}]
def hello():
    print("Hello")   # <-- difference
    x = 1            # <-- unchanged
    y = 2            # <-- unchanged
    return x + y     # <-- difference

z = hello()
\end{lstlisting}

\begin{lstlisting}[language=Python, caption={Rejected Code}]
def hello():
    print("Hi")      # <-- difference
    x = 1            # <-- unchanged
    y = 2            # <-- unchanged
    return x * y     # <-- difference

z = hello()
\end{lstlisting}

Prefix-Suffix-based method incorrectly marks both \texttt{x = 1} and \texttt{y = 2} as differences, while the LCS-based method correctly identifies only the two truly changed lines (\texttt{print("Hi")} and \texttt{return x * y}, providing a more precise grounding of the differences, enabling more targeted training.

\subsection{Different Negative Samples}

We chose the random negative selection method to better simulate the variety of error types in real-world scenarios, where some versions are severely flawed and others only slightly. This random slection approach enhances the diversity of negative samples and helps improve generalization.

\newpage
To validate this choice, we conduct additional experiments using only the first or last incorrect versions as negatives. The results in Table~\ref{select_negatives} show that our method consistently achieves superior performance on six benchmarks.

\subsection{Different Teacher Model}
We adopt DeepSeek-V3 (671B MoE LLM with 37B active parameters) as the teacher model and synthesize a total of 28k pairs with resource constraints. The results using Qwen2.5-Coder-7B as the student model are shown in Table~\ref{teacher_ablation}. Target-DPO demonstrates generalizable improvement when using different teachers for distillation.

\end{document}